\pdfoutput=1
\documentclass[preprints,article,accept,pdftex,moreauthors]{Definitions/mdpi}

\usepackage{xstring}

\usepackage{xstring}

\makeatletter
\AtBeginDocument{%
  \let\orig@includegraphics\includegraphics
  \renewcommand{\includegraphics}[2][]{%
    \IfSubStr{#2}{logo-mdpi}{}{%
      \IfSubStr{#2}{logo-updates}{}{%
        \orig@includegraphics[#1]{#2}%
      }%
    }%
  }%
}
\makeatother
\usepackage{amssymb}
\usepackage{algorithm}
\usepackage{algorithmic}
\usepackage{longtable}
\usepackage{array}
\usepackage{booktabs} 
\usepackage{amsmath}
\usepackage{color,soul}
\usepackage{url}
\usepackage{multirow} 
\usepackage[dvipsnames]{xcolor}
\usepackage{xcolor}
\sethlcolor{yellow}

\usepackage{comment}

\usepackage[justification=centering]{caption}
\firstpage{1} 
\pubvolume{1}
\issuenum{1}
\articlenumber{0}
\pubyear{2025}
\copyrightyear{2025}
\datereceived{ } 
\daterevised{ } 
\dateaccepted{ } 
\datepublished{ } 
\hreflink{https://doi.org/} 

\Title{\textbf{FARM}: Fine-Tuning Geospatial Foundation Models for Intra-Field Crop Yield Regression}

\Author{Shayan Nejadshamsi $^{1}$\orcidA{}, Yuanyuan Zhang $^{2}$, Shadi Zaki $^{2}$, Brock Porth $^{2}$ , Lysa Porth $^{2}$and Vahab Khoshdel $^{1}$\orcidA{}$*$}

\AuthorNames{Shayan Nejadshamsi, Yuanyuan Zhang and Vahab Khoshdel}

\isAPAStyle{%
       \AuthorCitation{Nejadshamsi, S., Zhang, Y., \& Khoshdel, V.}
         }{%
        \isChicagoStyle{%
        \AuthorCitation{Lastname, Firstname, Firstname Lastname, and Firstname Lastname.}
        }{
        \AuthorCitation{Lastname, F.; Lastname, F.; Lastname, F.}
        }
}

\address{%
$^{1}$ \quad University of Manitoba, Winnipeg, MB, Canada; shayan.nejadshamsi@umanitoba.ca\\
$^{2}$ \quad AIRM Consulting Ltd., Canada; yuanyuan.zhang@airmconsulting.com}

\corres{Correspondence: vahab.khoshdel@umanitoba.ca}

\abstract{Accurate and timely crop yield prediction is crucial for global food security and modern agricultural management. Traditional methods often lack the scalability and granularity required for precision farming. This paper introduces \textbf{FARM} (\textbf{F}ine-tuning \textbf{A}gricultural \textbf{R}egression \textbf{M}odels), a deep learning framework designed for high-resolution, intra-field canola yield prediction. FARM leverages a pre-trained, large-scale geospatial foundation model (Prithvi-EO-2.0-600M) and adapts it for a continuous regression task, transforming multi-temporal satellite imagery into dense, pixel-level (30 m) yield maps. Evaluated on a comprehensive dataset from the Canadian Prairies, FARM achieves a Root Mean Squared Error (RMSE) of $0.44$ and an $R^2$ of $0.81$. Using an independent high-resolution yield monitor dataset, we further show that fine-tuning FARM on limited ground-truth labels outperforms training the same architecture from scratch, confirming the benefit of pre-training on large, upsampled county-level data for data-scarce precision agriculture. These results represent improvement over baseline architectures like 3D-CNN and DeepYield, which highlight the effectiveness of fine-tuning foundation models for specialized agricultural applications. By providing a continuous, high-resolution output, FARM offers a more actionable tool for precision agriculture than conventional classification or county-level aggregation methods. This work validates a novel approach that bridges the gap between large-scale Earth observation and on-farm decision-making, offering a scalable solution for detailed agricultural monitoring.}

\keyword{Crop Yield Prediction; Precision Agriculture; Foundation Model; Satellite Sensing; Vision Transformer (ViT)} 

\begin{document}

\section{Introduction}
\label{introduction}

Accurate crop yield prediction is important for sustainable agricultural management, food and economic stability \cite{BASSO2019201,doi:10.1126/science.1204531}. As the global population grows \cite{UN_WPP_2024} and climate variability intensifies, producing reliable forecasts has become increasingly important for planning and adaptation \cite{lesk2016influence}. Robust yield estimates support decision-making across the food-supply chain, including government planning, farmers’ resource allocation and risk management \cite{shahhosseini2021coupling,fan2022gnn}.

Conventional prediction methods consist of field surveys and process-based crop models. Although they offer valuable insights, they are constrained by cost, scalability, and simplifying assumptions that often break down under diverse farming conditions \cite{khaki2020cnn,leng2019crop}. Expansion of Earth Observation (EO) remote-sensing technologies, like satellite and Unmanned Aerial Vehicles (UAVs) imagery, has transformed agricultural monitoring field by delivering continuous, high-quality observations of crop growth \cite{weiss2020remote,tahir2023application}. This data abundance has shifted yield prediction toward data-driven methods, where machine/deep learning methods capture the nonlinear, intricate properties \cite{oikonomidis2023deep}.

Traditional approaches relied on machine learning models like Random Forests using engineered vegetation indices \cite{mupangwa2020evaluating,kumar2024survey,sharma2023precision,pandey2024enhancing}, though these often lacked generalization across agro-climatic regions \cite{qomariyah2024applying}. Subsequent deep learning methods improved feature extraction via Convolutional Neural Networks (CNNs) \cite{nevavuori2019crop,khaki2020cnn} and captured temporal dynamics using Recurrent Neural Networks (RNNs) or attention mechanisms \cite{sun2019county,lin2020deepcropnet,albahli2022efficient}. However, these models are typically trained from scratch on limited datasets, potentially missing the broader Earth-system context \cite{weiss2020remote,tahir2023application}

Recently, a new computer vision field driven by Vision Transformers (ViTs) and geospatial foundation models has emerged. ViTs use self-attention to capture long-range dependencies within imagery, addressing CNNs’ tendency to focus on local patterns \cite{dosovitskiy2020image}. For instance, Barman et al. introduced ViT-SmartAgri and demonstrated that ViT architectures combined with smartphone-based applications can outperform traditional models like InceptionV3 in real-time plant disease detection. \cite{agronomy14020327} Similarly, Mehdipour et al. provided a survey highlighting that ViTs offer superior scalability and handling of long-range dependencies compared to CNNs, particularly in complex field environments \cite{mehdipour2025visiontransformersprecisionagriculture}. Furthermore, hybrid approaches and lightweight ViTs are being actively explored to reduce computational costs while maintaining high accuracy in precision agriculture tasks \cite{s23156949}. Large-scale foundation models, like IBM–NASA’s Prithvi-EO-2.0-600M, are pre-trained on global satellite archives and provide generalizable representations that can be fine-tuned for downstream tasks, including agricultural prediction \cite{szwarcman2024prithvi}. Together, these advances point toward more scalable and transferable frameworks for data-driven crop yield forecasting.

Despite progress, several gaps remain. First, to the best of our knowledge, many existing models are trained from scratch on task-specific agricultural image sets that are relatively small \cite{nevavuori2019crop,nevavuori2020crop,sun2019county}, limiting generalization and forgoing the Earth-system knowledge embedded in pre-trained foundations. Second, most deep learning models target coarse county-level yield, which is useful for regional planning but insufficient for capturing intra-field variability—defined here as the localized yield differences within a single field boundary caused by soil heterogeneity and management zones—which is required by farmers and operational decision-makers. Third, higher-resolution efforts often frame yield as a classification problem (discretized categories), which cannot fully represent the continuous nature of productivity or mixed pixels. Fourth, many crop-yield models are treated as “black boxes” with limited interpretability. To earn trust from decision-makers, models should pair accuracy with insights into the agronomic processes driving predictions. For example, this may include the relative importance of specific growth stages or spectral signatures.
To address these limitations, we introduce the \textbf{FARM} framework 
(\textbf{F}ine-tuning \textbf{A}gricultural \textbf{R}egression \textbf{M}odels). FARM adapts a pre-trained ViT-based encoder for intra-field crop yield regression, providing a more detailed and flexible tool for mapping agricultural productivity. Building FARM on a foundation model allows us to leverage its rich, pre-trained understanding of the spatio-temporal structure of satellite data for our canola yield-prediction task.

We hypothesize that fine-tuning a pre-trained geospatial foundation model (Prithvi-EO) will result in lower error rates (lower RMSE) compared to similar architectures trained from scratch, due to the model's ability to transfer learned representations of complex Earth-surface dynamics. We further hypothesize that training FARM on a large, Upsampled county-level supervised dataset with upsampled county-level labels can produce a robust base model that transfers effectively to truly intra-field yield prediction when fine-tuned on a much smaller set of high-resolution yield monitor labels. The objectives of this study are to:

\begin{itemize}
    \item Adapt and fine-tune a geospatial ViT-based foundation model for an intra-field level yield prediction and show the efficiency of using foundation models for crop yield prediction tasks in comparison with other state-of-the-art approaches. 
    \item Design a pixel-wise regression architecture that frames yield prediction as a continuous task at high spatial resolution, enabling the model to estimate a precise yield value for every image pixel.
    \item Assess interpretability by quantifying and visualizing embedded attention mechanisms to identify influential phenological stages and spectral signatures that drive the model's predictions.
    \item Demonstrate, using an independent high-resolution yield monitor dataset, that fine-tuning FARM provides a stronger performance than training the same architecture from scratch, establishing a practical transfer-learning pathway from regional monitoring to precision agriculture.

\end{itemize}

As a key contribution to crop-yield prediction, we pioneer the application of a large-scale geospatial foundation model (Prithvi-EO-2.0-600M) to yield forecasting in the Canadian Prairies, showing how pre-trained models can be effectively adapted for specialized agricultural tasks. While we focus on canola, the framework is designed to be adaptable to other crops.

The remainder of this paper is structured as follows. Section 2 details the methodology, including datasets and the FARM architecture. Section 3 describes the study experimental setup model configurations, and training objectives. Section 4 presents and discusses results, including performance metrics and baseline comparisons. Sections 5 and 6 provide a summary of findings and directions for future research.

\section{Methodology}
\label{methodology}

\subsection{Dataset and Preprocessing}
\label{sec:s}

This section details the data sources and data preprocessing stages of this study.

\subsubsection{Data Source and Characteristics}

\begin{figure}[H]
\centering
\includegraphics[width=\textwidth]{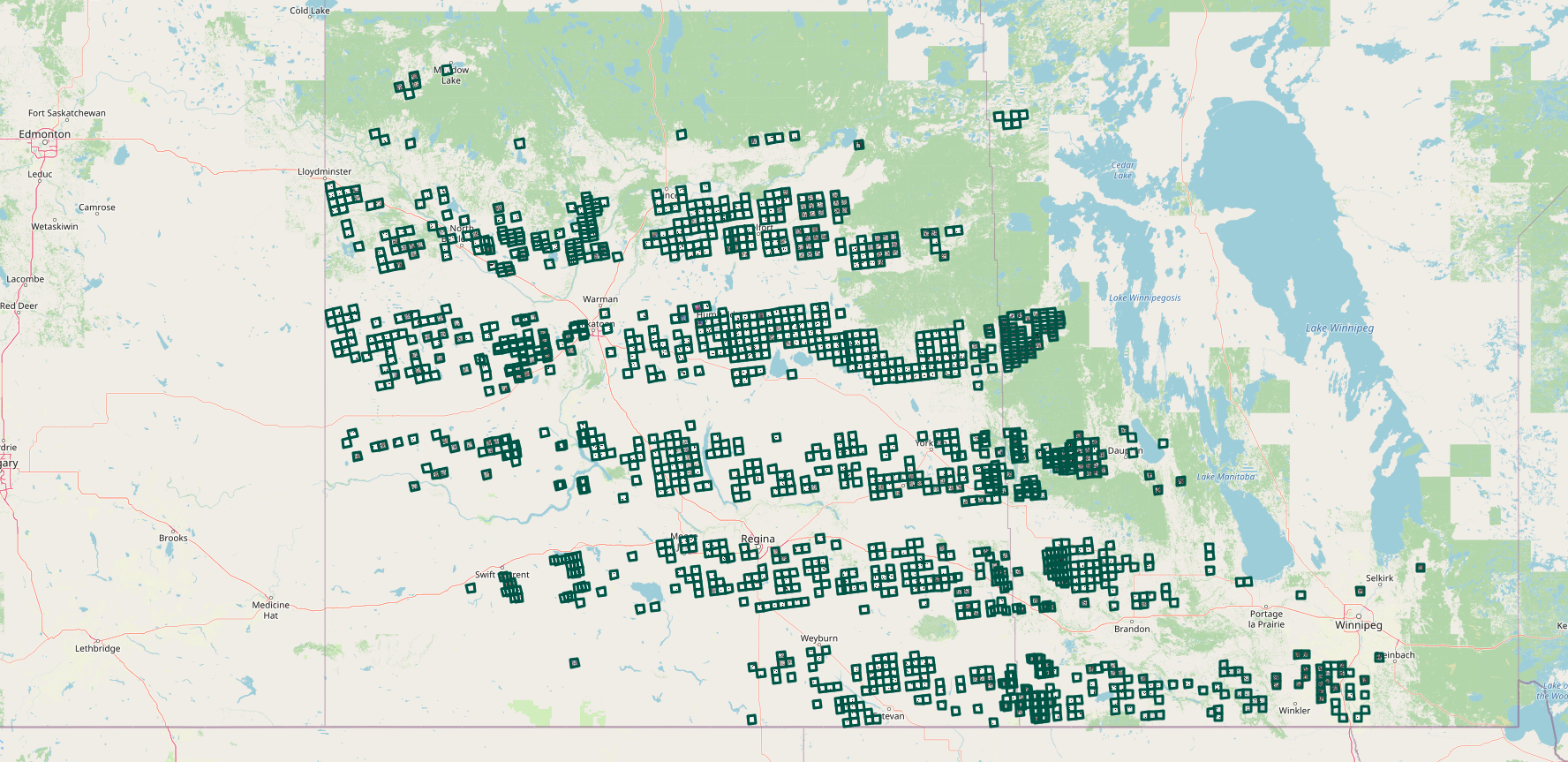}
\tiny
\caption{Geographic distribution of the image chips used in this study across canola-growing regions of Saskatchewan and Manitoba in the Canadian Prairies. Each small area represents a unique geographical area for which multi-temporal satellite imagery and ground-truth yield data were collected.}
\label{fig:fig_1}
\end{figure}

This study use a multi-temporal image and crop yield dataset. The dataset integrates high-resolution satellite image chips with corresponding up-sampled ground-truth yield maps at the same resolution. The geographical focus of this dataset is the Canadian Prairies, a major global region for canola production. The spatial distribution of the data samples in our region of interest is shown in Figure \ref{fig:fig_1}.

\textbf{Satellite Imagery:} The model input includes multi-temporal satellite imagery that cover the entire growing season of canola from May to September. The dataset source is Sentinel-2 satellite imagery processed into the Harmonized Landsat Sentinel-2 (HLS) format. It contains spatiotemporal samples, each corresponding to a different geographical image chip located primarily in Saskatchewan and Manitoba, Canada. For each chip sample, the data includes a time-series of images. Each image chip includes six spectral bands: Blue, Green, Red, Near-Infrared (Narrow), Short-Wave Infrared 1 (SWIR 1), and Short-Wave Infrared 2 (SWIR 2). For each chip sample, the data includes a time-series of images structured as five distinct temporal steps, corresponding to monthly composite images from May to September for the years 2018, 2019, 2020, 2022 and 2023. While Sentinel-2 provides a 5-day revisit time, frequent cloud cover and atmospheric interference in the Canadian Prairies often result in data gaps that prevent the use of the full temporal resolution. To ensure high data quality, we first filtered the image collection to exclude individual scenes where cloud cover exceeded 10\%. Following this filtration, monthly compositing was selected to mitigate these environmental obstructions, ensuring spatially less noisy inputs. This aggregation strategy also maintains a consistent temporal dimension for the model input tensor, balancing computational efficiency with the need to capture the primary phenological stages of canola growth. This temporal resolution is designed to capture key phenological stages of canola growth, from early vegetative growth in May through peak flowering in July to maturity and senescence in September, which are important for accurate yield estimation.

\textbf{Ground-truth Yield Data:} 
For the main dataset used to train and validate FARM, we construct pixel-level yield maps by upsampling the low-resolution county-level yield values to the image resolution using bilinear interpolation. Although these labels are defined at 30 m resolution and provide intra-field variation at the pixel level, they should be regarded as upsampled county-level supervised targets rather than true point-level yield measurements. We later evaluate how well a model trained on these upsampled county-level labels can recover real spatial variance in Section~\ref{sec:highres_experiments}. This regression-based formulation allows the model to learn fine-grained crop yield variations within and across fields, capturing the effects of localized soil conditions, water availability, and different management practices. 

\textbf{Additional High-Resolution Dataset for Independent Validation:}
In addition to the upsampled county-level supervised labels, we also incorporate a separate dataset containing native high-resolution (10 m) yield monitor measurements collected across multiple fields in the Canadian Prairies (2013–2024). Importantly, this dataset shares the \emph{same multi-temporal Sentinel-2 HLS input configuration} described above; the only difference lies in how the pixel-level labels are constructed: (i) upsampled county-level yields for large-scale training, versus (ii) truly measured intra-field yield values for independent validation. The high-resolution yield monitor dataset is \textit{never used for training} the FARM model; it serves exclusively as an external test set to assess the model’s ability to generalize to real intra-field spatial variability. The corresponding experimental protocol and results are presented in Section~\ref{sec:highres_experiments}.

\begin{figure}[H]
\centering
\includegraphics[width=\textwidth]{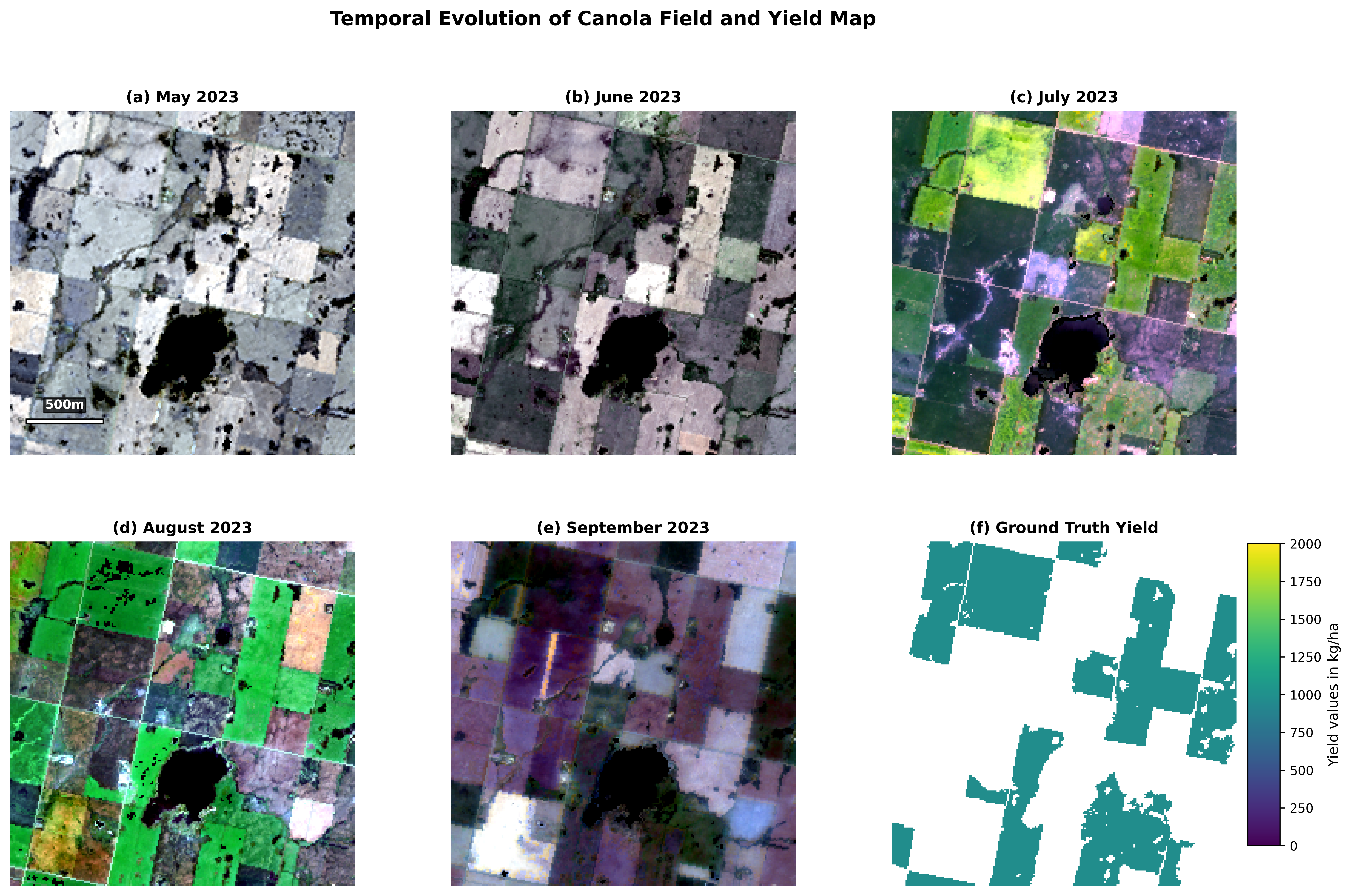}
\tiny
\caption{Temporal evolution of canola field and corresponding ground-truth yield map in a sample image chip. Panels (a)-(e) show RGB composite images derived from Harmonized Landsat Sentinel-2 (HLS) data from May to September 2023, representing the phenological progression of canola from early vegetative growth through flowering to senescence. Panel (f) displays the ground-truth yield map with values in kg/ha.}
\label{fig:fig_2}
\end{figure}

Figure \ref{fig:fig_2} provides a visual representation of a data sample from our dataset for the year 2023, showing both the temporal dimension of the input data and the corresponding ground-truth. The multi-temporal satellite imagery encompasses the entire growing season of canola development across the growing season.

\subsubsection{Data Structuring}

The multi-temporal image data are structured into a tensor format as the input to the ViT encoder, which is designed to process both the spectral and temporal dimensions simultaneously. Both dimensions are flattened along the channel and time dimensions to create a single multi-channel tensor. Specifically, an input with $T$ time steps and $C$ channels per time step is transformed into a tensor of shape $[C \times T, H, W]$, where $H$ and $W$ are the height and width of the image chip. In our case, the input for a single sample (image chip) ($224 \times 224$ pixels) is a sequence of five images, each with six spectral bands, resulting in a final model input tensor with 30 channels. The corresponding ground-truth is a single-channel tensor of shape $[1, H, W]$ containing the continuous yield values. This structure allows the initial patch embedding layer of the ViT to treat temporal steps as distinct channels, enabling the subsequent self-attention layers to model complex interdependencies across both space and time.

For our time-series prediction problem, we use a temporal hold-out validation strategy, where we evaluate the model's generalization to unseen growing seasons. The training set comprises 6,079 image chips, while the validation set contains 1,749 chips.

\subsubsection{Preprocessing and Data Augmentation}

\textbf{Normalization:} To standardize the feature space, each of the six spectral bands was normalized independently using a pre-calculated channel-wise mean and standard deviation derived from the training dataset. The channel-wise mean and standard deviation values are shown in Table \ref{tab:tab_1}:

\begin{table}[H]
    \centering
    \caption{Channel-wise mean and standard deviation}
    \small
    \label{tab:tab_1}
    \begin{tabularx}{\textwidth}{LCC} \toprule
        \textbf{Spectral bands}  & \textbf{Means}     &   \textbf{Standard Deviations} \\ \midrule
        BLUE        & 493.94    &   250.38  \\ 
        GREEN       & 832.45    &   265.75 \\ 
        RED         & 901.06    &   481.92 \\ 
        NIR NARROW  & 2927.87   &   1038.83 \\ 
        SWIR 1      & 2427.47   &   855.02 \\ 
        SWIR 2      & 1658.56   &   855.37 \\
        \bottomrule
    \end{tabularx}
\end{table}

\textbf{Data Augmentation:} On-the-fly data augmentation was applied during model training to increase the diversity of the training set and mitigate overfitting. Geometric transformations, including random horizontal and vertical flips, were applied with a probability of 20\%. While a probability of 50\% is standard in general computer vision tasks, we opted for a more conservative 20\% threshold. This decision was made to introduce sufficient regularization to prevent overfitting while preserving the inherent directional properties of the satellite imagery, such as sun-sensor geometry and crop row orientation, which can be distorted by excessive geometric transformation. These augmentations create new training examples without altering the yield labels, encouraging the model to learn more powerful and invariant feature representations. 

\subsection{Model Architecture}
\label{sec:s}

\begin{figure}[H]
\centering
\includegraphics[width=\textwidth]{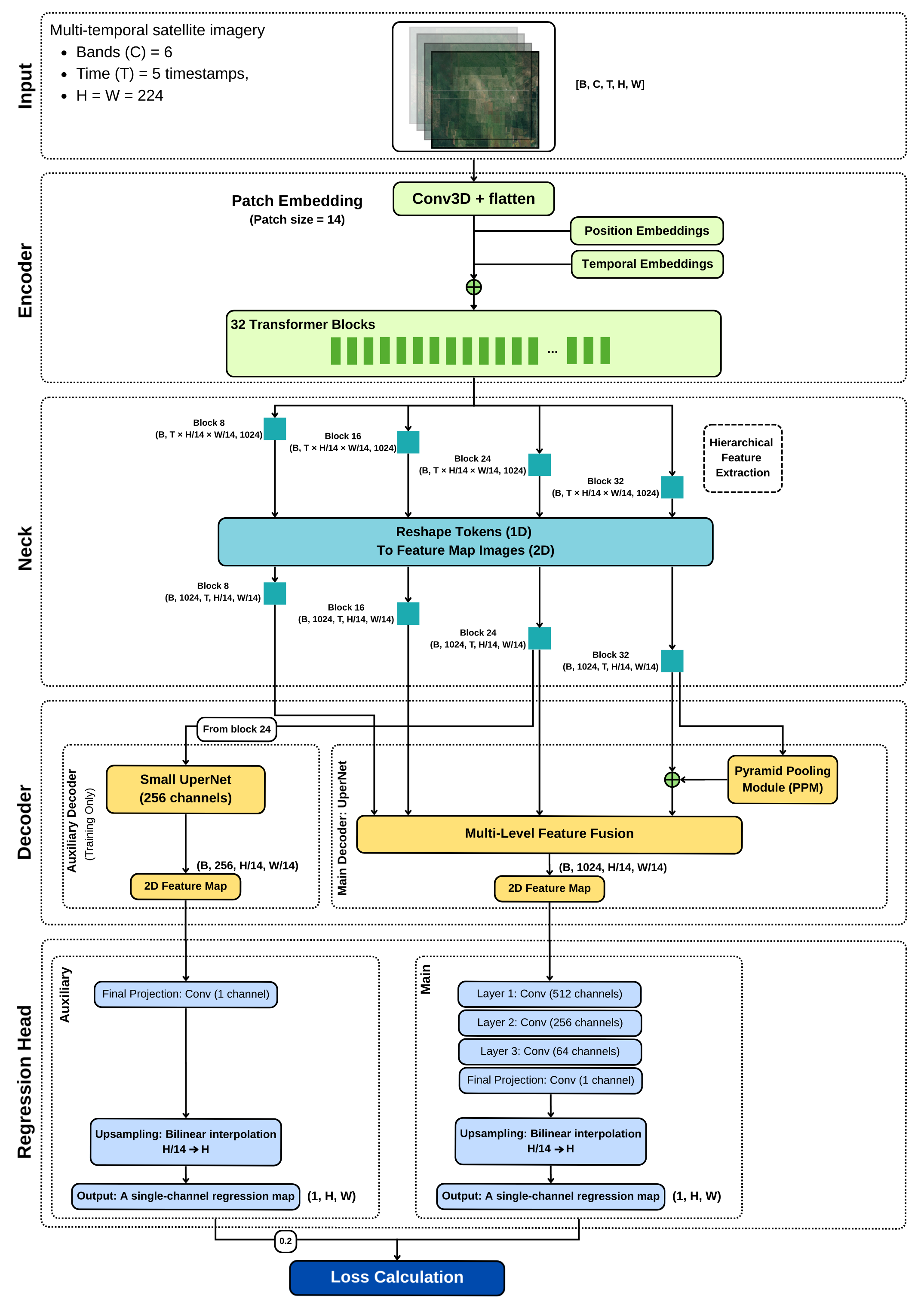}
\caption{FARM Architecture}
\label{fig:fig_3}
\end{figure}

Given the multi-temporal nature of the HLS dataset described above, we selected a model architecture specifically capable of handling sequential geospatial features. Our proposed intra-field level crop yield regression model (FARM), is designed as an encoder-decoder architecture suitable for dense prediction tasks where a high-resolution output map must be generated from a high-dimensional input. Our model uses the feature extraction capabilities of a pre-trained Prithvi ViT as its encoder, complemented by a decoder and a specialized regression head. The overall architecture of the FARM model is illustrated in Figure \ref{fig:fig_3}. This architecture is designed to transform multi-temporal satellite imagery into a dense yield map, directly learning the relationships between spectral-temporal patterns from the data. In the following, the details of each main module are described.

\vspace{0.5cm}

\subsubsection{Encoder: Prithvi-EO-2.0-600M Vision Transformer}

This primary function of this module is the feature extraction task. This large-scale geospatial foundation model is based on the Vision Transformer (ViT) architecture and pre-trained on a large global dataset of satellite image sequences using masked autoencoding—a self-supervised learning technique where the model learns to reconstruct hidden portions of an input image to understand underlying spatial structures. It provides a powerful and generalized understanding of Earth observation data. The process of encoding is as follows:

\textbf{Input and Patch Embedding:} The model input is in the form 30-channel multi-temporal image tensor of shape $[B, 30, H, W]$. This input is first divided into a grid of non-overlapping 14 $\times$ 14 patches. A 2D convolution is then applied to flatten each patch and project it into an embedding space.

\textbf{Spatio-temporal Embeddings:} To provide important context, learnable embeddings are added to the patch tokens. For example, positional embeddings are used to include spatial information about the location of each patch within the original image. Temporal and location embeddings are used to model the geographic and seasonal context of the imagery, which allows the model to differentiate between images taken at different times and locations.

\textbf{Transformer Blocks:} The sequence of embedded tokens is then processed by a series of 32 transformer blocks. Each block is composed of three main sub-layers: 1) \textit{Multi-Head Self-Attention (MHSA):} This is the core mechanism of the ViT. It allows the model to weigh the importance of all other patches when representing a given patch, thereby capturing global contextual information across the entire image. 2) \textit{Feed-Forward MLP:} A standard multi-layer perceptron, which is applied to each token independently. 3) \textit{LayerNorm and Residual Connections:} Layer Normalization is applied before each sub-layer, and residual connections are used around each sub-layer to facilitate stable training of the deep network.

During the fine-tuning process for our specific task, the entire pre-trained encoder is kept unfrozen, allowing its weights to be updated and adapted. The output of the encoder is a sequence of processed tokens, where each token contains a context-aware representation of its corresponding image patch.   

\subsubsection{Decoder: UperNet for Feature Reconstruction}

To reconstruct a spatially detailed output from the abstract token representations generated by the ViT encoder, we employ a UperNet decoder. UperNet is a powerful and widely used architecture for semantic segmentation that excels at fusing features from multiple scales. It aggregates features from four selected layers of the transformer encoder (specifically layers 8, 16, 24, and 32), which represent different levels of semantic abstraction. Given that the encoder consists of 32 transformer blocks, selecting these specific layers divides the network into four evenly spaced stages. This configuration is a standard heuristic for adapting isotropic ViT architectures to pyramidal decoders, ensuring that the reconstructed feature pyramid captures a balanced progression from low-level spatial patterns to high-level semantic representations. The decoding process involves several key stages: 1) \textit{Reshaping:} The selected token outputs from the transformer are first reshaped back into 2D feature maps. 2) \textit{Feature Pyramid Network (FPN):} The multi-scale feature maps are then progressively merged using an FPN. Lateral connections from the transformer's feature maps are combined with upsampled features from deeper layers. This process systematically integrates high-resolution, low-level features with low-resolution, high-level semantic features. 3) \textit{Pyramid Scene Parsing (PSP):} In this stage, the output of the FPN is processed by a PSP module. This module applies pooling at multiple scales to aggregate scene-level context, to capture a more holistic understanding of the feature map. 4) \textit{Bottleneck:} The final fused features from the PSP module are passed through a 3 $\times$ 3 convolutional bottleneck, resulting in a single, rich feature map ready for the final prediction head.

\subsubsection{Convolutional Regression Head}

The final component of our architecture is a specialized regression head that maps the feature map from the UperNet decoder to a single-channel as a yield prediction map.This regression head uses a series of three 3 $\times$ 3 convolutions to gradually reduce the channel dimensionality to single channel, while refining the spatial features. Each intermediate convolutional layer is followed by a Batch Normalization, a ReLU activation function to ensure stable training and introduce non-linearity. The final output tensor has shape of $[B, 1, H, W]$, where each pixel value represents the predicted canola yield. 

\section{Experiments}

\subsection{Training Objective and Implementation}
\label{sec:s}

To train our model, we define an objective function and utilize a training pipeline with modern optimization and regularization techniques. The entire framework is implemented to ensure efficient, stable, and reproducible training.

The model's goal is to predict a continuous yield value, $\hat{y}_{i,j}$, for each pixel $(i,j)$ in an input image, such that it closely matches the corresponding ground-truth yield, $y_{i,j}$. To quantify the discrepancy between predicted and true yield maps, we evaluate two  loss functions.

The primary loss function used in our main experiments is the Mean Squared Error (MSE) loss. It is a standard and effective choice for regression tasks, calculated as the average of the squared differences between the predicted and actual yield values over all pixels in a batch. For a single predicted yield map $\hat{Y}$ and a ground-truth map $Y$, both of size $H \times W$, the MSE is defined as:

\begin{equation}
\mathcal{L}_{MSE}(Y,\hat{Y}) = \frac{1}{H \times W} \sum_{i=1}^{H} \sum_{j=1}^{W} (y_{i,j} - \hat{y}_{i,j})^2
\end{equation}

MSE loss is differentiable and penalizes larger errors more heavily due to the squaring term, which encourages the model to avoid significant deviations.

To assess the model's robustness to potential outliers in the ground-truth data, we also conducted experiments using the Huber Loss. The Huber loss is a piecewise function that provides a compromise between the sensitivity of MSE and the robustness of Mean Absolute Error (MAE). It behaves quadratically for small errors but linearly for large errors, which reduces the influence of outliers that might otherwise dominate the gradient during training. This makes the Huber loss less sensitive to anomalous yield values in the dataset, often leading to better generalization.

\begin{equation}
\mathcal{L}_{\delta}(y, \hat{y}) = 
\begin{cases} 
\frac{1}{2}(y - \hat{y})^2, & \text{if } |y - \hat{y}| \le \delta \\ 
\delta|y - \hat{y}| - \frac{1}{2}\delta^2, & \text{otherwise} \end{cases}
\end{equation}

where $\delta$ is a tunable hyperparameter that defines the threshold at which the function transitions from quadratic to linear. 

To facilitate the stable training of our architecture, we employ a deep supervision strategy through the use of an auxiliary head. This auxiliary head is attached to an intermediate layer of the UperNet decoder and provides an additional, coarser-grained prediction output. The purpose of this auxiliary head is to inject an additional gradient signal directly into the middle of the network during backpropagation. This helps to mitigate the vanishing gradient problem, which can be a challenge in very deep models, and encourages the intermediate layers to learn more discriminative features. The auxiliary head has its own smaller UperNet decoder and a 1-channel regression output. Its loss, calculated using the same primary loss function (e.g., MSE), is added to the total loss of the main head, weighted by a factor of 0.2. The total loss for the model is therefore:

\begin{equation}
\mathcal{L}_{total} = \mathcal{L}_{main} + 0.2 \times \mathcal{L}_{auxiliary}
\end{equation}

This deep supervision technique helps to regularize the model and often leads to faster convergence and improved final performance.

\subsection{Evaluation Metrics}
\label{sec:s}

To evaluate the model's performance, we utilize four standard regression metrics: Mean Absolute Error (MAE) \cite{hodson2022root}, Root Mean Squared Error (RMSE) \cite{hodson2022root}, Coefficient of Determination ($R^2$) \cite{ozer1985correlation}, and the Pearson Correlation Coefficient \cite{benesty2009pearson}. While RMSE and MAE quantify the magnitude of prediction error, $R^2$ and the Pearson Correlation Coefficient assess the model's ability to explain yield variability and capture linear trends in the data.

\subsection{Implementation Details}
\label{sec:s}

The entire framework was implemented using the PyTorch deep learning library and streamlined with PyTorch Lightning for organized and reproducible training. To ensure experimental reproducibility, the random seed was fixed to 0. We initialized the encoder using the pre-trained Prithvi EO V2 600M TL weights and trained the full architecture for 120 epochs with a batch size of 8. We utilized the AdamW optimizer, which decouples weight decay from gradient-based updates, applying a weight decay of 0.1 to improve generalization. To dynamically adjust the learning rate, we employed a Cosine Annealing scheduler preceded by a linear warm-up period of 20 epochs; the learning rate started at $5 \times 10^{-6}$ and decayed to a minimum of $1 \times 10^{-8}$. During training, data augmentation was applied on-the-fly, including random horizontal and vertical flips with a probability of 0.2 and Gaussian noise injection with a probability of 0.4 to enhance robustness against sensor noise. The UperNet decoder was configured with 1024 channels to effectively fuse multi-scale features, while the regression head utilized a dropout rate of 0.1 to mitigate overfitting. To accelerate training and reduce GPU memory consumption, we utilized bfloat16 (bf16) mixed-precision training. The experiments were conducted on a high-performance computing node equipped with NVIDIA GPUs with 48 GB of memory.

\subsection{Baseline}
\label{sec:s}

To further contextualize the performance of our foundation model-based approach, we compare it against other state-of-the-art baseline architectures for spatio-temporal prediction tasks. To ensure a fair comparison, we adopted the 3D-CNN and DeepYield architectures proposed in~\cite{nevavuori2020crop} and~\cite{GAVAHI2021115511} respectively, and trained them from scratch using our canola yield dataset under identical experimental settings. These model were selected because, as reported in~\cite{nevavuori2020crop,GAVAHI2021115511}, these architectures have already demonstrated statistically significant performance gains over traditional machine learning baselines (including Random Forest, LASSO, and SVM) and other deep learning approaches. Consequently, they serve as the most reliable benchmark for evaluating the advancements offered by FARM. However, since the original study focused on a different crop type, a direct comparison would not have been equitable. Therefore, retraining the models on our dataset enabled a consistent and crop-specific performance evaluation. The 3D-CNN architecture follows an encoder–decoder design that organizes multi-temporal satellite image data into a unified spatiotemporal framework and employs 3D convolutional kernels to jointly learn spatial and temporal dependencies. DeepYield frames the architecture as a combination of 3D-CNN and ConvLSTM.

\section{Results}
\label{results}

This section presents the evaluation results of our proposed intra-field level crop yield regression (FARM) framework. First, we detail the quantitative performance, followed by a qualitative analysis of the generated yield values. Subsequently, we compare our proposed model against other baselines to highlight the efficiency of our proposed model architecture in predicting continuous yield values. Finally, we present an analysis of the model's interpretability through its attention weights from temporal and spectral perspectives to explore the key components involved in the final predictions.

\subsection{Quantitative Evaluation}
\label{sec:s}

Table~\ref{tab:tab_2} presents a comparison of intra-field regression performance for three training configurations (loss functions)—MSE, Huber, and MSE + Aux—on the validation set. As described in the Methodology section, the model was trained under three distinct strategies: (i) using only the Mean Squared Error (MSE) loss, (ii) using only the Huber loss, and (iii) using the MSE loss combined with an auxiliary head for deep supervision.

The results are reported in both standardized form (as used during training) and de-standardized into common agricultural units (kg/ha and bu/ac) for practical interpretability. The model trained with the Huber loss achieved a slightly lower RMSE of 0.4677 and a higher $R^2$ of 0.7852 compared to the MSE-only model. This modest improvement indicates that while outliers are present in the dataset, they do not strongly influence the regression outcomes; nonetheless, employing a robust loss such as Huber yields a small gain in generalization.

The best model (FARM) achieved an $R^2$ of $0.8105$, showing that it successfully explains over $81\%$ of the variance in pixel-level canola yield within the validation dataset. The high Pearson Correlation of $0.9003$ demonstrates a strong linear relationship between the predicted and ground-truth yield values, confirming that the model's predictions are consistently aligned with the actual yield trends. 

\begin{table}[H]
\centering
\small
\caption{A comparison of intra-field level regression performance for three training configuration (MSE, Huber, and MSE+Aux) on the validation set.}

\begin{tabularx}{\textwidth}{LCCCC} \toprule
 \multicolumn{2}{c}{} & \multicolumn{3}{c}{\textbf{Training configuration}}\\
\hline
 & \textbf{Metrics} & \textbf{MSE} & \textbf{Huber} & \textbf{MSE+Aux} \\
\midrule
\multirow{4}{*}{\textbf{RMSE}}  & standardized & 0.4782 & 0.4677 & 0.4368\\
                                & kg/ac & 92.76 & 90.63 & 84.54 \\
                                & kg/ha & 229.33  & 224.02 &208.79\\
                                & bu/ac & 4.09 & 3.99 &3.73\\
\hline
\multirow{3}{*}{\textbf{MAE}} & standardized & 0.3615 & 0.3545 & 0.3317\\
                                & kg/ac & 70.15 & 68.71 & 64.16 \\
                                & kg/ha & 173.31 & 169.79 &158.47 \\
                                & bu/ac & 3.09 & 3.03 &2.83\\
\hline
\textbf{R²} &           -        & 0.7727 & 0.7852 &0.8105 \\
\hline
\textbf{Pearson correlation} & - & 0.8791 & 0.8861 &0.9003\\
\bottomrule
\end{tabularx}
\label{tab:tab_2}
\end{table}

\begin{figure}[H]
\centering
\subfloat[\centering]{\includegraphics[width=7cm]{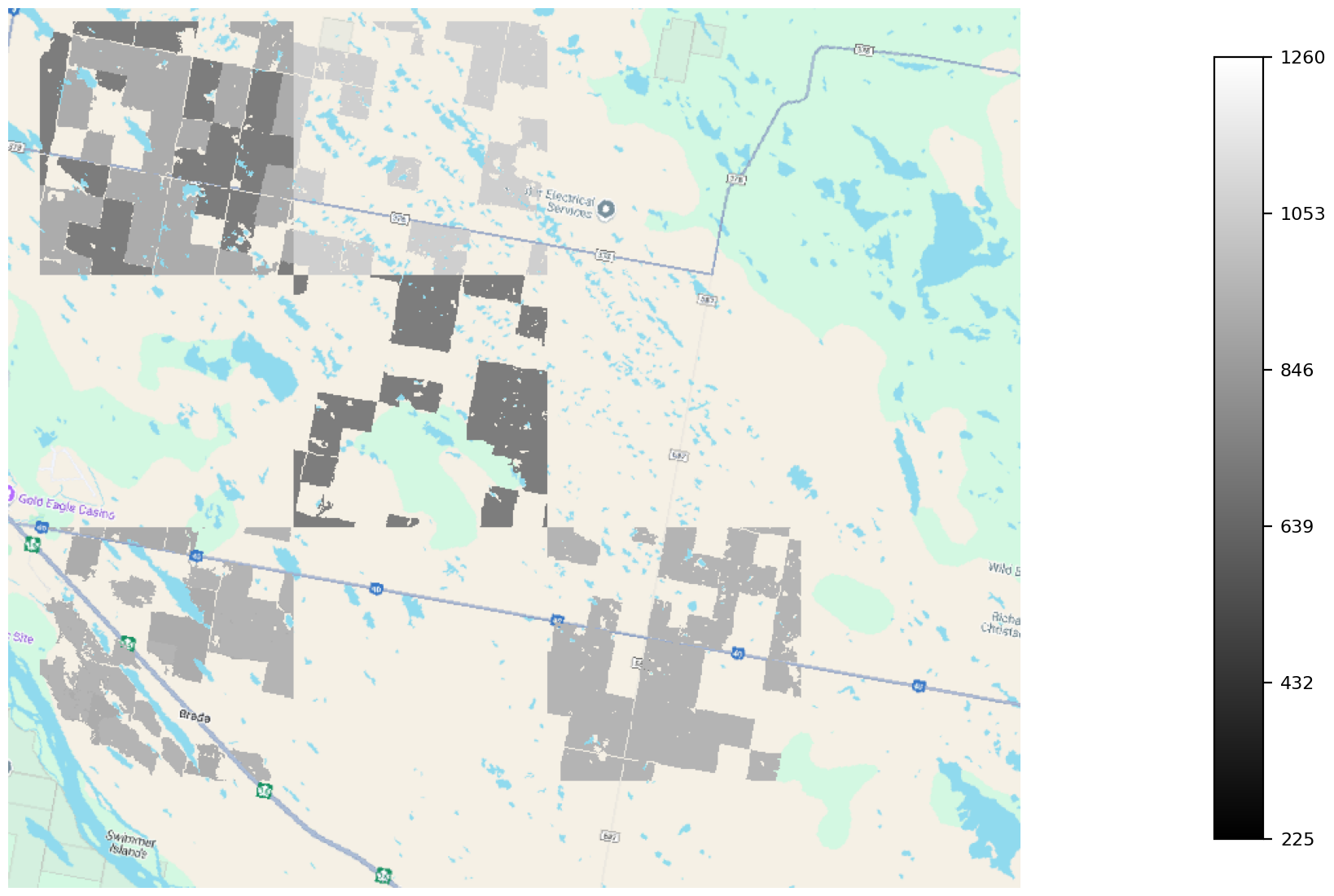}}
\subfloat[\centering]{\includegraphics[width=7cm]{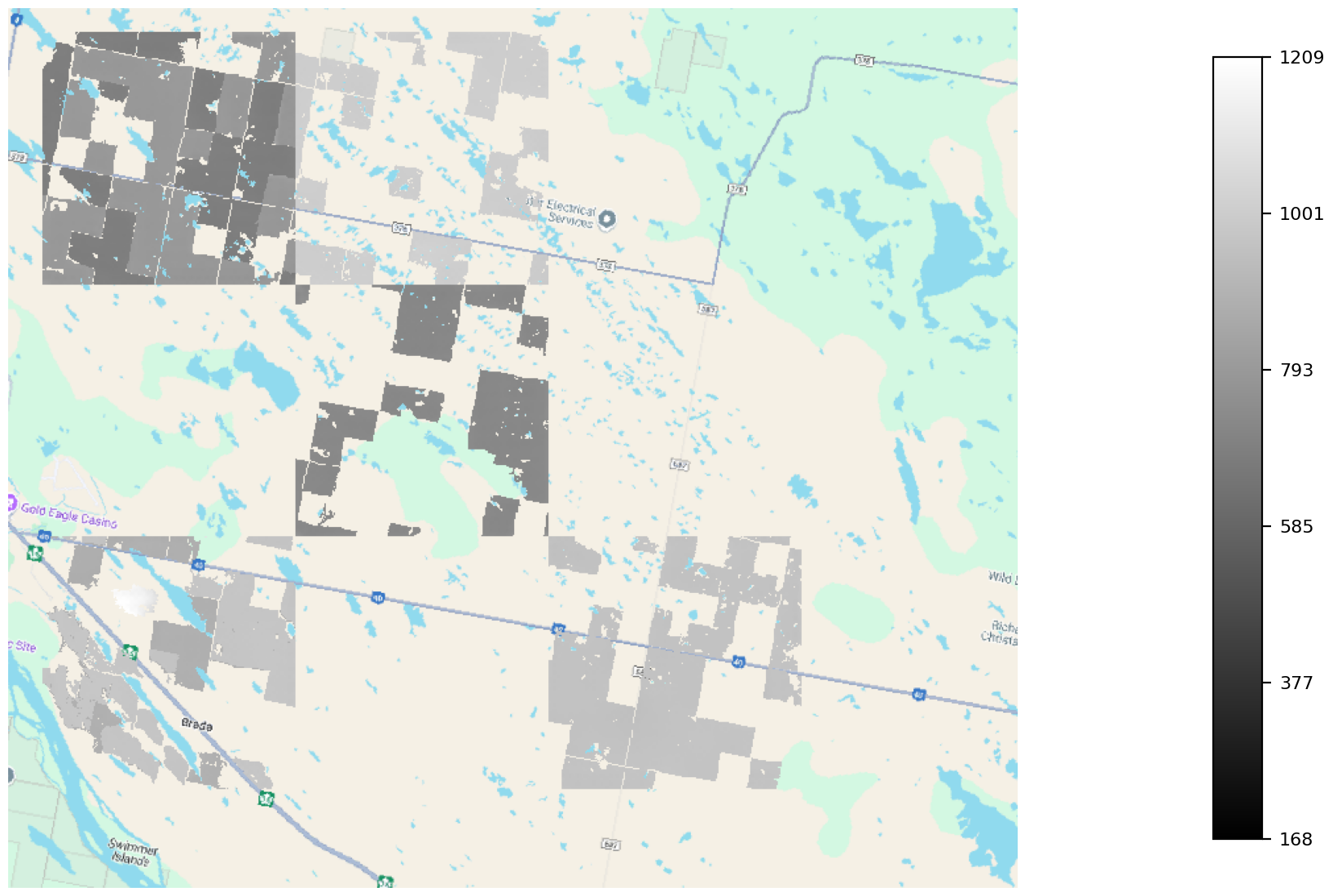}} \\
\subfloat[\centering]{\includegraphics[width=7cm]{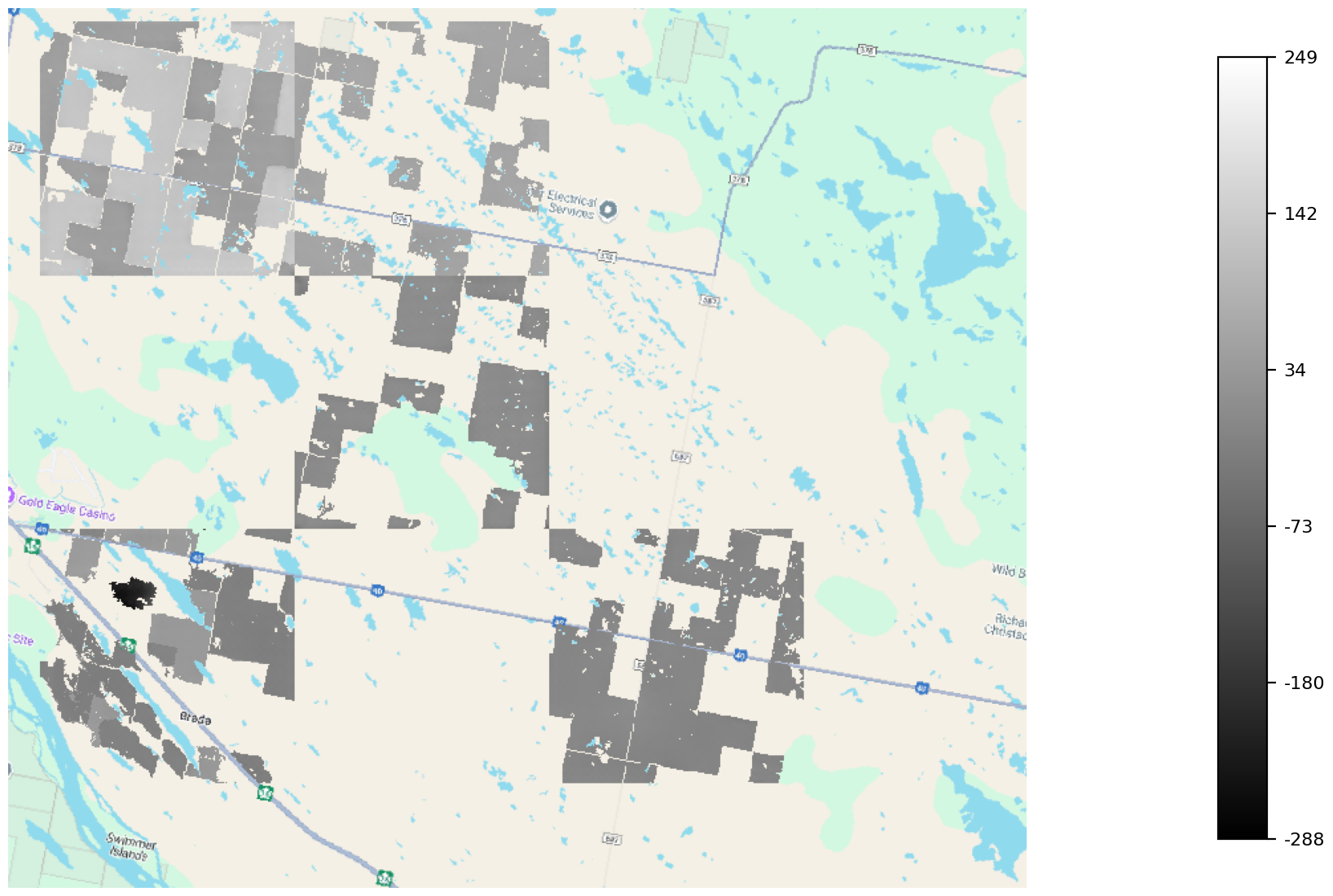}}
\caption{(\textbf{a}) Ground-truth yield map. (\textbf{b}) Model-predicted yield map. (\textbf{c}) Map of residuals, showing spatial error patterns}
\label{fig:fig_4}
\end{figure}

Figure \ref{fig:fig_4} presents side-by-side comparison of predicted yield values with ground-truth for an arbitrary location in our region of interest. The predicted yield map (Figure \ref{fig:fig_4} (b)) successfully captures the fine-grained, intra-field variability present in the ground-truth map (Figure \ref{fig:fig_4} (a)). It also includes a map of the residuals (\ref{fig:fig_4} (c)), which illustrates the spatial patterns of prediction errors, showing areas where the model demonstrates higher or lower accuracy.

\subsection{Qualitative Evaluation}
\label{sec:s}

\begin{figure}[H]
\centering
\includegraphics[width=\textwidth]{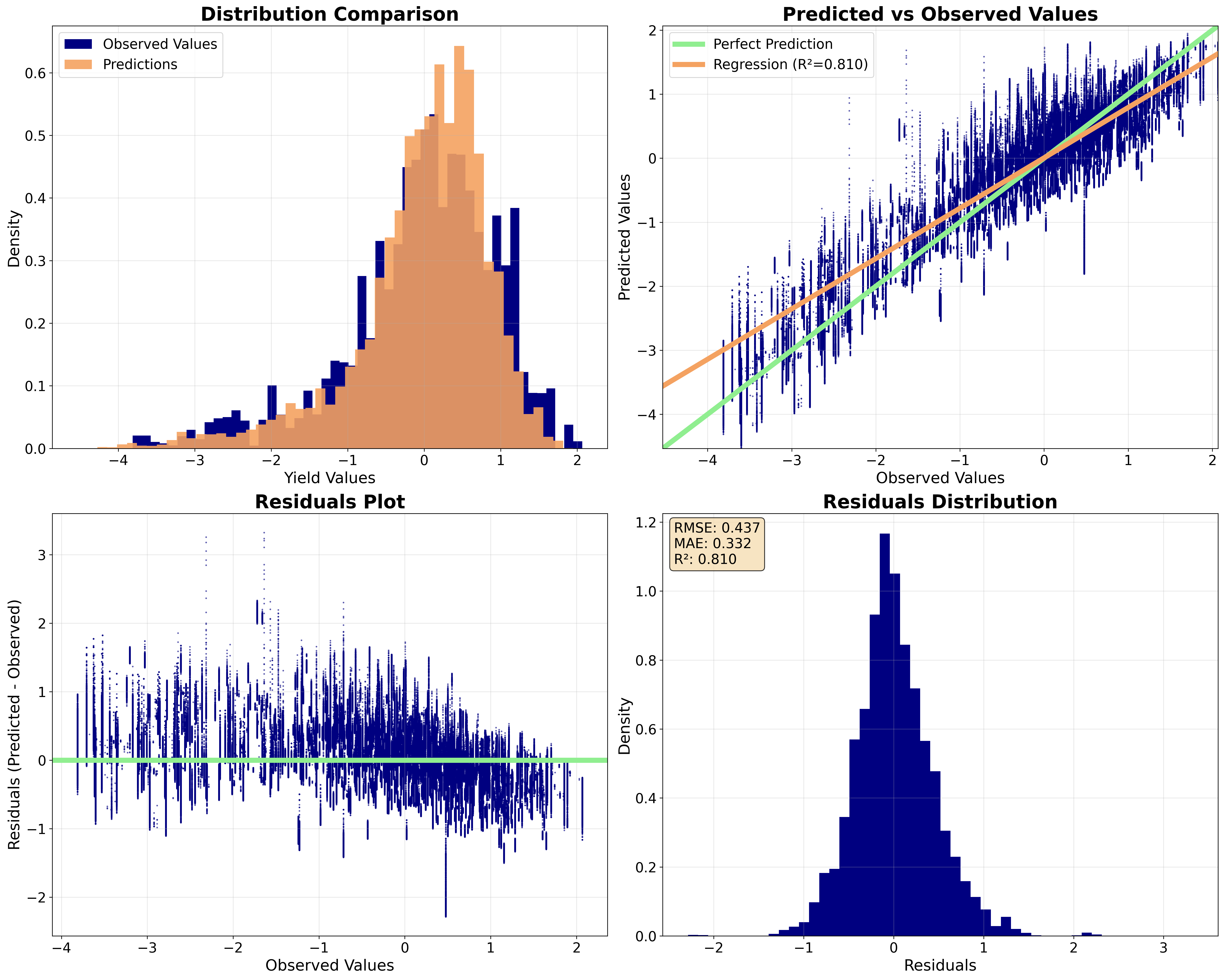}
\caption{Qualitative performance analysis (a) Distribution of prediction and ground-truth values (b and c) Scatter plot of predicted vs. true pixel values. (d) Histogram of prediction residuals.}
\label{fig:fig_5}
\end{figure}

To visually assess the model's spatial prediction capabilities, we compared the predicted yield maps against the ground-truth. Figure \ref{fig:fig_5} shows this for a representative sample from the validation set. The scatter plot of predicted versus true values (Figure \ref{fig:fig_5} (b and c)) shows a tight clustering of points along the identity line. This confirms the high correlation reported in Table \ref{tab:tab_2}. The distribution of prediction errors is centered around zero and is close to a normal distribution (Figure \ref{fig:fig_5} (d)), which indicates that the model does not have a significant systematic bias.

\subsection{Comparison with baselines}
\label{sec:s}

\textbf{Comparison with Regression-based baseline:}

To validate the effectiveness of adapting a large-scale foundation Prithvi-EO-2.0 model for this task, we compared our final model (FARM) against the state-of-the-art baselines and results are presented in Table 3.

The results clearly indicate that our FARM model, significantly outperforms the 3D-CNN and DeepYield models across all standard regression metrics. This higher performance can be attributed to two key factors. First, the Prithvi-EO-2.0-600M encoder has been pre-trained on a massive and diverse dataset of global satellite imagery. This provides our model with a rich, generalized understanding of vegetation dynamics, phenology, and land surface patterns that a model trained from scratch on a specific agricultural dataset cannot easily acquire.

\begin{table}[H]
    \centering
    \caption{Model performance comparison with regression-based baseline}
    \label{tab:tab_3}
    \begin{tabularx}{\textwidth}{LCCCC}
    \toprule
        Model       & RMSE        & MAE       &$R^2$    & Pearson Coefficient \\ \midrule
        3D-CNN \cite{nevavuori2019crop}  & 1.2852     & 0.9721   & 0.6248   & 0.6455 \\
        DeepYield \cite{GAVAHI2021115511} & 1.0322     & 0.8714   & 0.7323   & 0.7642 \\
        \textbf{FARM}   & \textbf{0.4368}     & \textbf{0.3317}   & \textbf{0.8105}   & \textbf{0.9003} \\ 
        \bottomrule
    \end{tabularx}
\end{table}
 Second, the Vision Transformer architecture, with its self-attention mechanism, is inherently capable of capturing long-range spatial dependencies across an entire image chip. This may allow it to better model the contextual factors influencing yield variability than the localized receptive fields of the CNN-based encoder in the baseline models. These results support our hypothesis that fine-tuning large geospatial foundation models is a superior strategy for complex, dense prediction tasks like intra-field level crop yield estimation. Due to differences in crop type relative to prior studies (which did not focus on canola), a direct comparison with all previously proposed baselines was not appropriate. Therefore, we restrict our baseline evaluation to 3D-CNN and DeepYield, which are reported as the strongest-performing models in those works \cite{nevavuori2020crop,GAVAHI2021115511}

\subsection{Validation on High-Resolution Ground Truth Data}
\label{sec:highres_experiments}

To address the limitations associated with training on upsampled county-level yield data, we performed a series of experiments using a distinct dataset containing high spatial resolution (10 m) ground-truth yield data. This dataset covers limited numbers of fields in Canadian Prairies, collected during 2013 to 2024 growing seasons. Unlike the primary training set, these labels were not upsampled but represent true localized yield values.

We conducted three experiments to assess the performance of the FARM architecture and the transferability of the features learned from the upsampled data.

\subsubsection{Experiment 1: Direct Inference (Zero-Shot Application)}

\begin{figure}[H]
\centering
\includegraphics[width=\textwidth]{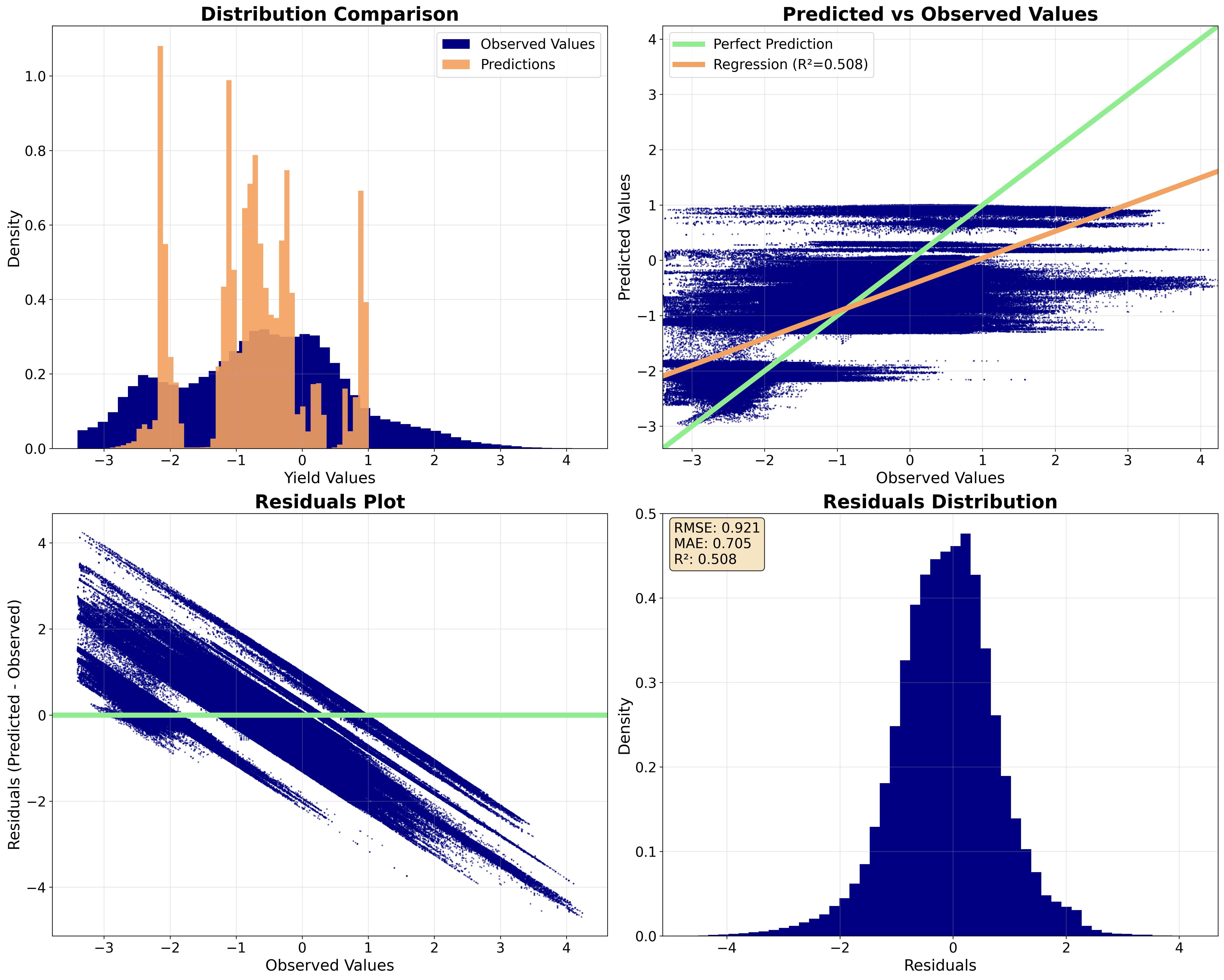}
\caption{Qualitative performance analysis for Experiment 1: Direct Inference (Zero-Shot Application)}
\label{fig:fig_10}
\end{figure}

We applied the FARM model (originally trained on upsampled county-level labels) directly to the new high-resolution dataset without any weight updates. The objective is to determine if the model learned generalized spectral-yield relationships or merely memorized the smoothed regional trends. All imagery originally at 10 m spatial resolution is upsampled to 5 m in order to match the 224-dimensional input specification required by the county-level FARM model checkpoint. This transformation doubles the spatial dimensions (112 → 224) through interpolation applied directly to the raw imagery. Each 10 m pixel is subdivided into four 5 m pixels using standard resampling methods(bilinear interpolation), and no boundary-filling strategies (such as padding with mean values) are applied. The resulting 5 m rasters preserve the original spatial extent while producing inputs that are fully compatible with the higher-resolution model workflow. The model achieved an RMSE of 0.921 and an $R^2$ of 0.508. Figure \ref{fig:fig_10} shows the qualitative and quantitative presentation of this experiment.

\subsubsection{Experiment 2: Fine-Tuning on High-Resolution Ground-truth Data}

\begin{figure}[H]
\centering
\includegraphics[width=\textwidth]{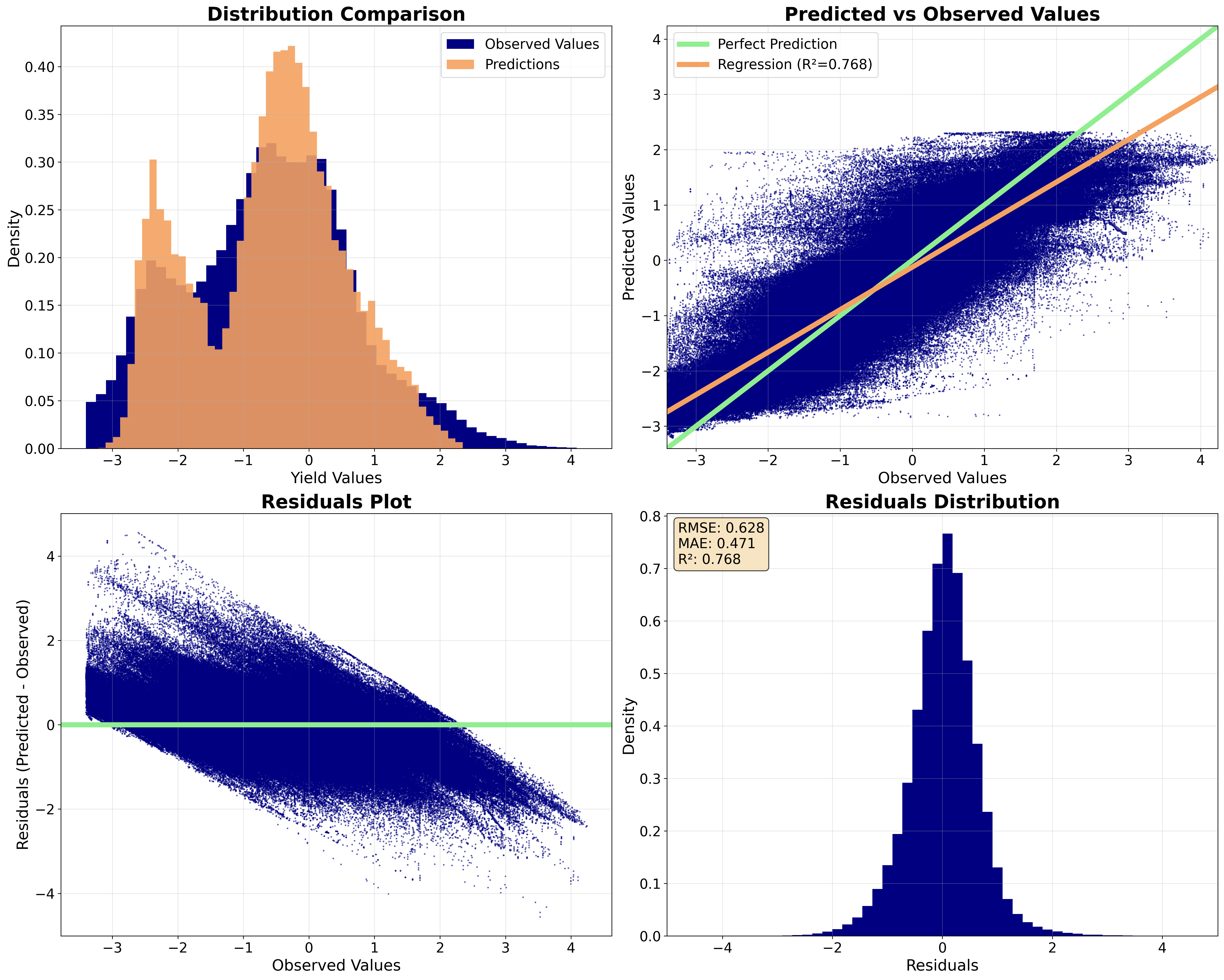}
\caption{Qualitative performance analysis for Experiment 2: Fine-Tuning on High-Resolution Ground-truth Data}
\label{fig:fig_11}
\end{figure}

We utilized the pre-trained FARM model (from the main study) and fine-tuned it on the high-resolution dataset. The objective is  to test if the representations learned from the Foundation Model and refined on upsampled data serve as a strong initialization for high resolution ground-truth data and high precision agriculture tasks. The fine-tuned model achieved an RMSE of 0.628 and an $R^2$ of 0.768. Figure \ref{fig:fig_11} shows the qualitative and quantitative presentation of this experiment.

\subsubsection{Experiment 3: Training on High-Resolution Ground-truth Data}

\begin{figure}[H]
\centering
\includegraphics[width=\textwidth]{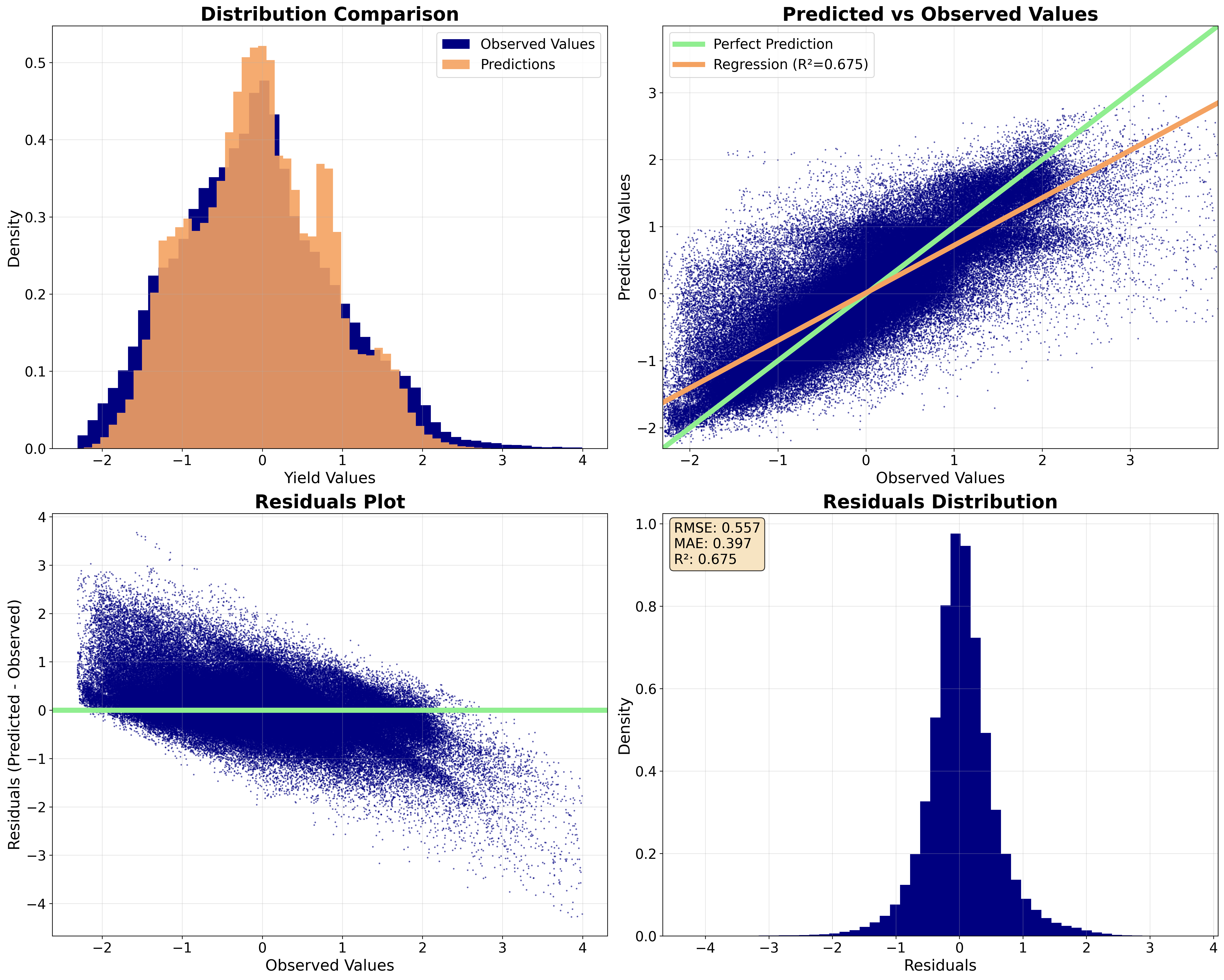}
\caption{Qualitative performance analysis for Experiment 3: Training on High-Resolution Ground-truth Data}
\label{fig:fig_12}
\end{figure}

We initialized the FARM architecture with the standard Prithvi-EO-2.0-600M weights to be trained on high resolution ground-truth data from scratch. Some changes happened in traiing pipline for example using Heteroscedastic loss instead of MSE/Huber loss and also adding brightness and contrast augmentation on top of augmentations used in the model based on county-level labels to the input during training. This model achieved an RMSE of 0.557 and an $R^2$ of 0.675. Figure \ref{fig:fig_12} shows the qualitative and quantitative presentation of this experiment.

Table \ref{tab5} summarizes the performance metrics across the original and three experimental setups. These results confirm that the features learned from the upsampled county-level data are transferable and agronomically meaningful. In Experiment 1, the Zero-Shot application yielded an $R^2$ of 0.508, indicating that the model captures fundamental yield drivers even without seeing native high-resolution labels during training. However, Experiment 2 achieved the highest explanatory power ($R^2=0.768$) by leveraging the pre-trained weights from the main study (trained on upsampled data) and fine-tuning them on the high-resolution dataset. Notably, this approach outperformed Experiment 3 ($R^2=0.675$), where the model was initialized randomly and trained from scratch on the high-resolution data. This performance gap underscores the challenge of training deep architectures on limited datasets; the high-resolution dataset, while precise, contained few samples for the model to fully converge on generalized features from scratch. The results suggest the robustness of the architecture but its potential is constrained by data volume in Experiment 3, and it would likely achieve higher accuracy if trained on a larger corpus of high-resolution imagery. Consequently, using massive volumes of upsampled county-level data to create a foundation serves as a critical initialization step for tasks where granular ground truth is scarce.

\begin{table}[H] 
\caption{Quantitative performance comparison across three validation setups using high-resolution ground-truth data.\label{tab5}}
\begin{tabularx}{\textwidth}{CCCC}
    \toprule
\textbf{}    & \textbf{RMSE}    & \textbf{MAE}  & \textbf{$R^2$}\\
\midrule
FARM - Using Up-sampled Ground-truth Data  & 0.4368           & 0.3317  & 0.8105\\ \hline
Experiment 1: Direct Inference (Zero-Shot Application)      & 0.921           & 0.705 & 0.508\\ \hline
Experiment 2: Fine-Tuning on High-Resolution Ground-truth & 0.628    & 0.471           & 0.768 \\ \hline
Experiment 3: Training on High-Resolution Ground-truth & 0.557        & 0.397        & 0.675 \\
\bottomrule
\end{tabularx}
\end{table}


\subsection{Interpretability Analysis}
\label{sec:s}

 The interpretability of deep learning models in agriculture is of paramount importance, as it enables stakeholders to understand the underlying factors driving model predictions and fosters trust in AI-assisted decision-making. Beyond achieving high predictive accuracy, interpretable models allow agronomists, producers, and policymakers to gain meaningful insights into how temporal patterns, such as crop growth dynamics and seasonal variability, and other environmental or management-related features contribute to yield outcomes. This interpretability analysis was performed for both the temporal and spectral (channel) aspects to understand which phenological stages and what spectral information the model considers most influential for predicting canola yield.

\begin{figure}[H]
\centering
\includegraphics[width=\textwidth]{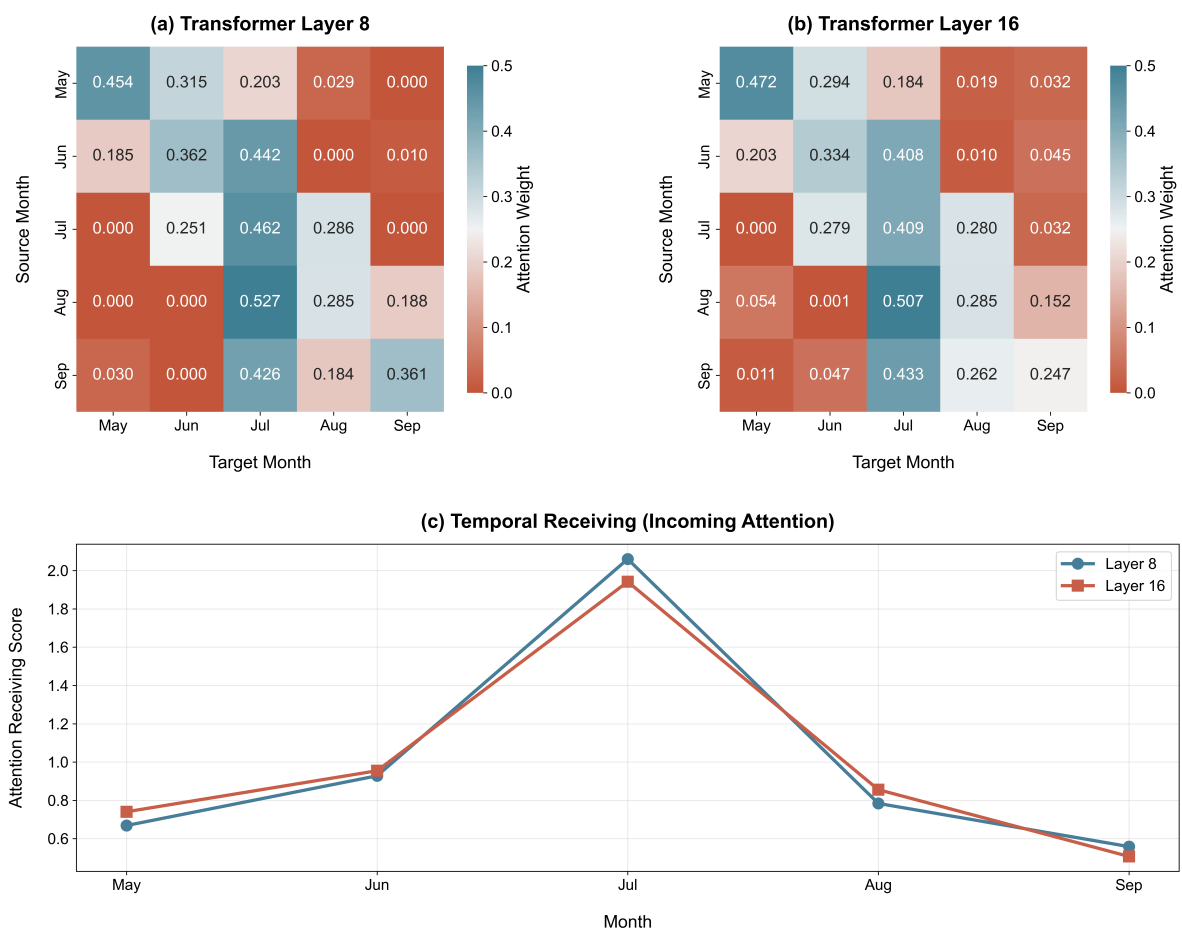}
\tiny
\caption{Temporal attention patterns from the FARM model for canola yield prediction. The heatmaps display the attention weights from (a) Transformer Layer 8 and (b) Transformer Layer 16. The x-axis represents the target month, and the y-axis represents the source month, with darker shades indicating higher attention weights. (c) The line graph illustrates the temporal receiving score (incoming attention) for each month.}
\label{fig:fig_6}
\end{figure}

By analyzing the attention matrices from key transformer layers, we quantified the focus the model places on each of the five monthly time steps (May through September). The results reveal a compelling and highly interpretable pattern. In the earlier layers of the transformer, such as Layer 8 (Figure \ref{fig:fig_6} (a)), the model exhibits a strong, localized attention pattern, where each time step primarily attends to itself and its immediate neighbors. For instance, the July time step shows a dominant self-attention score, indicating the model is learning to consolidate information within this critical period. As we progress to deeper layers, such as Layer 16 (Figure \ref{fig:fig_6} (b)), the attention mechanism evolves to capture more complex, long-range temporal dependencies. Here, the model consistently assigns the highest importance to the mid-season months, with July emerging as the most influential time step, receiving significant attention from both earlier and later periods. This is quantitatively evidenced by the higher aggregate attention scores for July and August across multiple attention heads. This pattern aligns perfectly with established crop physiology; for canola, the flowering and early pod-filling stages occurring in July are paramount for yield determination, as they directly influence seed set and development. The model's ability to autonomously identify and prioritize this peak growing season, without any explicit phenological guidance, underscores its capacity to learn biologically salient features directly from the spectral-temporal data. Furthermore, the attention maps show that while the late-season month of September receives less direct attention, it still plays a contextual role, likely helping the model discern maturity and senescence patterns. This temporal interpretability not only builds trust in the model's predictions but also scientifically validates that the fine-tuned foundation model successfully internalizes the growth dynamics of canola, effectively focusing its analytical power on the most agronomically decisive periods of the growing season.

\begin{figure}[H]
\centering
\includegraphics[width=\textwidth]{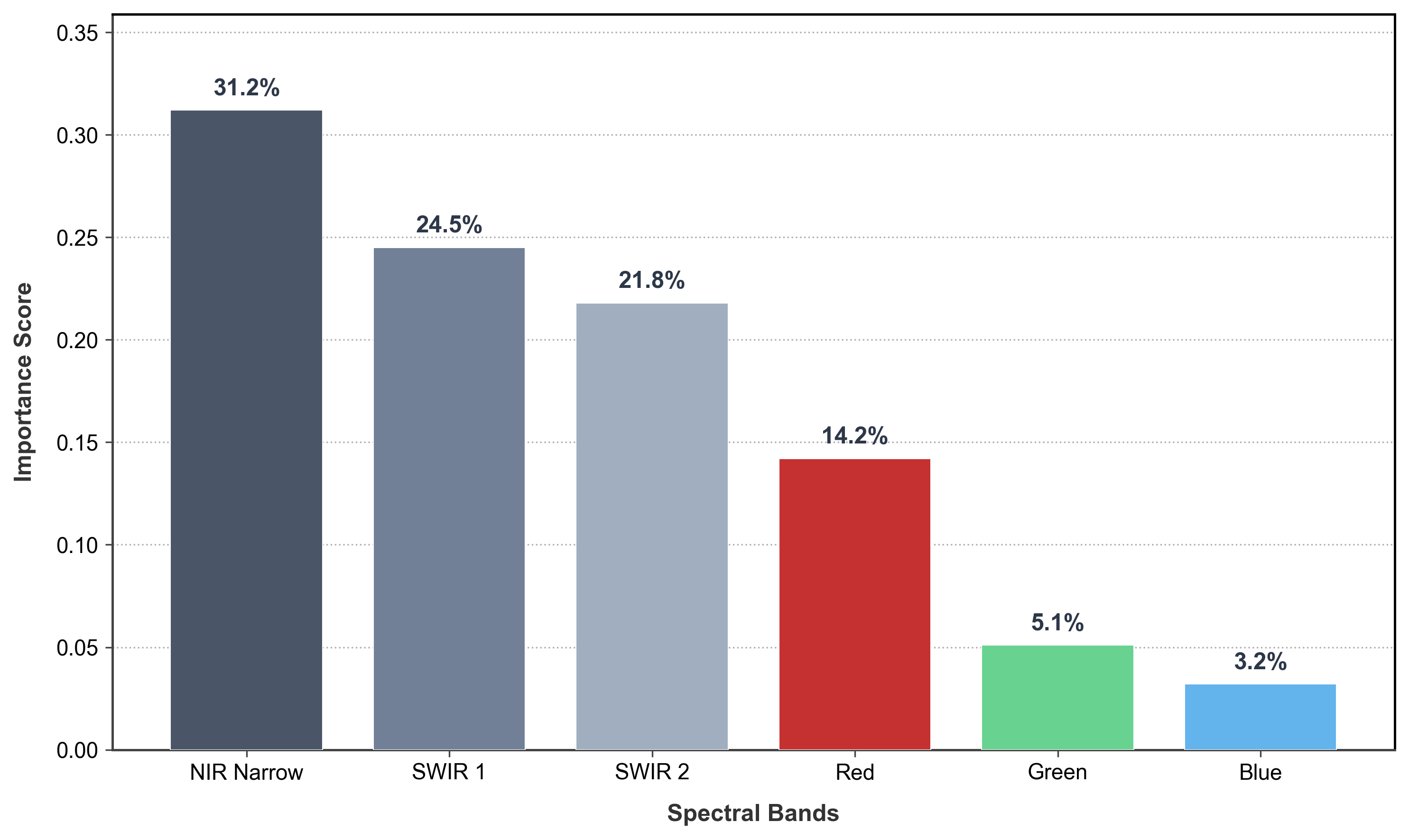}
\tiny
\caption{Spectral Importance Analysis: The relative importance score for each of the six input spectral bands as learned by the FARM model.}
\label{fig:fig_7}
\end{figure}

Additionally, it is important to understand which image spectral bands the model uses to make its predictions. We conducted a channel-wise analysis by analyzing the magnitude of the learned patch embedding weights corresponding to each of the six input spectral bands across all time steps. The results of this technique show the relative importance the model assigns to each band at the very initial stage of processing, which provides insight into which spectral features are considered most informative. According to Figure \ref{fig:fig_7}, the analysis reveals that the Near-Infrared (NIR) and Short-Wave Infrared (SWIR 1 and SWIR 2) bands received the highest importance scores. This observation is consistent with established remote sensing principles for vegetation monitoring \cite{holzman2021relationship,wang2008sensitivity,hunt2007remote}. NIR, which is the basis of calculating vegetation indices like NDVI is directly related to plant growth status. SWIR bands are the next significant bands. These bands are sensitive to moisture content of vegetation and soil, which highly influence crop yield. The Red visible band is placed at the next rank due to its' contributing role as the other component in NDVI. The Blue and Green bands were found to be the least influential. 

\section{Discussion}
\label{discussion}

The results of our proposed FARM framework show a noticeable improvement in canola yield prediction with a significant MAE of 2.83 bushels per acre. These gains underscore the value of incorporating a foundation model compared to hybrid spatiotemporal models like 3D-CNN and DeepYield. While 3D-CNNs, DeepYield and similar hybrid architectures are efficient for learning spatiotemporal patterns, their knowledge is fundamentally limited to the canola-specific patterns they extract during training. By contrast, the superior performance of FARM can be directly attributed to its fine-tuning of the Prithvi-EO-2.0-600M foundation model. By leveraging a model pre-trained on a massive and diverse archive of global satellite imagery \cite{szwarcman2024prithvi}, FARM benefits from a rich, generalized understanding of vegetation dynamics, atmospheric conditions, and land surface phenology. This large and transferable knowledge base allows the model to interpret the nuances of canola growth in more detail, leading to higher accuracy in yield prediction. Our work thus shifts from training task-specific models toward fine-tuning generalist foundation models for specialized agricultural applications, enabling more scalable solutions.

In addition, this study reformulates the yield prediction as a high-resolution intra-field regression task, which marks a critical methodological advance for precision agriculture. While recent state-of-the-art frameworks, such as the GNN-RNN approach by \cite{fan2022gnn} and DeepCropNet \cite{lin2020deepcropnet}, have demonstrated high accuracy in capturing spatial-temporal dependencies, they primarily target county-level or regional aggregation. Although these models are valuable for macro-scale food supply planning, their resolution is insufficient for capturing the granular, intra-field heterogeneity driven by local soil and management conditions. Historically, high-resolution yield prediction has often been framed as a low-resolution classification problem, discretizing yields into a finite set of categories. Such a classification approach imposes artificial boundaries on a continuous biological process, resulting in significant loss of granularity and unrealistic sharp transitions in predicted productivity. Our regression-based FARM model, in contrast, generates continuous yield maps that represent the smooth spatial heterogeneity present within and across agricultural fields. These high-resolution maps can directly inform decision-makers and precision-farming operations.

A key consideration in interpreting these results is the generation of pixel-level training labels through the upsampling of county-level data. While this approach allows for large-scale training where granular data is scarce, it raises questions regarding the model's ability to capture true intra-field heterogeneity versus simply smoothing regional averages. To validate the model’s capacity for high-resolution prediction, we conducted a supplementary analysis using a separate dataset containing ground-truth yield data at native 30 m resolution. As detailed in Section~\ref{sec:highres_experiments}, we evaluated the model under three conditions: direct inference, fine-tuning, and training from scratch. The results demonstrate that the FARM model, initially trained on upsampled county-level data, successfully learns generalized spectral-yield relationships rather than simply memorizing smoothed regional averages. While direct inference (Zero-Shot) on high-resolution data showed moderate performance ($R^2 = 0.51$), fine-tuning the pre-trained model on true high-resolution ground truth significantly improved accuracy ($R^2 = 0.77$). Crucially, this fine-tuned model outperformed a version of the architecture trained from scratch solely on high-resolution data ($R^2 = 0.68$). The lower performance observed in the model trained from scratch is primarily attributed to the limited number of images in the high-resolution dataset. Thus, our transfer-learning strategy currently offers the most viable solution for bridging the gap between data scarcity and precise intra-field prediction.

Another contribution of this work lies in the interpretability analysis of the FARM model. Temporal attention analysis (Figure \ref{fig:fig_7}) shows that the model autonomously learns to assign the highest importance to mid-season months, with July consistently emerging as the most influential time step. This learned behavior aligns with the established crop physiology of canola \cite{https://doi.org/10.2135/cropsci2003.1358}. The period spanning July and August corresponds to the critical flowering and pod-filling stages, during which the plant’s sensitivity to environmental conditions is at its peak and the primary determinants of final seed yield are established. The model’s ability to identify and prioritize this critical time window from raw spectral–temporal data indicates that it is capturing fundamental drivers of crop development.

Furthermore, the analysis of which spectral bands the model leverages provides deeper insight into its decision-making process. The model assigns the highest importance to the Near-Infrared (NIR) and Short-Wave Infrared (SWIR 1 and SWIR 2) bands, which aligns with principles in remote sensing for vegetation monitoring \cite{holzman2021relationship,wang2008sensitivity,hunt2007remote}. Healthy vegetation strongly reflects NIR light, which makes this band important for plant analysis.[2][4][5] The SWIR bands are sensitive to the moisture content in both vegetation and soil, which are effective markers that significantly influence crop yield. The model's reliance on these specific bands demonstrates that it has learned to focus on the spectral signatures most indicative of plant health and water status, which are key drivers of canola yield.

While the current findings demonstrate the efficacy of the proposed approach, it is important to acknowledge the limitations of the study, which define specific paths for future research. The FARM model was trained and validated on canola within the context of the Canadian Prairies. To validate broader applicability, future work will focus on extending the model through multi-crop fine-tuning strategies. Specifically, we aim to adapt the architecture for wheat yield prediction, investigating whether the shared geospatial representations in the foundation model can be leveraged to enhance performance across distinct crop types with varying phenological cycles. Although the model shows high performance using multi-temporal imagery alone, there is significant opportunity to enhance its predictive power by integrating meteorological and soil data, thereby providing a more holistic view of agricultural ecosystems.

To translate FARM from a research capability to a deployable precision agriculture tool, the generated pixel-level yield maps must integrate directly with Farm Management Information Systems (FMIS). In a practical workflow, these yield predictions would serve as the foundational layer for Variable Rate Application (VRA) prescriptions. By identifying high-yielding and low-yielding zones early in the season, agronomists could adjust nitrogen inputs or seeding rates dynamically to maximize Return of Interest (ROI) per acre. Additionally, these maps can guide targeted scouting, allowing growers to physically inspect anomaly zones detected by the model rather than relying on random field sampling.

However, several gaps remain before commercial deployment. First, the latency between satellite acquisition (HLS data availability) and inference needs to be minimized to ensure decisions can be made within tight operational windows. Second, future iterations must incorporate uncertainty quantification, providing users with a confidence interval alongside yield estimates to build trust in automated decision-making.

\section{Conclusions}

In this research, we introduced FARM, a foundation model for intra-field crop yield prediction, and demonstrated its effectiveness on canola in the Canadian Prairies. Our work establishes the significant advantages of fine-tuning a large-scale, pre-trained geospatial foundation model, Prithvi-EO-2.0-600M, for specialized agricultural tasks, which shows superior performance over models trained from scratch. In FARM, we also formulated the yield prediction as a continuous, pixel-level regression task. We addressed the limitations of low-resolution discrete classification. This approach generates high-resolution, continuous yield maps that capture the granular, intra-field variability which is important for precision agriculture applications. Furthermore, our analysis of the model's temporal attention mechanisms showed that FARM's output for crop yield prediction prioritizes the most critical phenological stages for canola yield, which are aligned with the current literature. This confirms its ability to learn agronomically meaningful patterns. Overall, the findings position foundation models as a new technology for advancing data-driven crop yield prediction.

\funding{This research was funded by DIGITAL – Canada’s Global Innovation Cluster for Digital Technologies (project “AI-Enabled Enterprise Risk Management for the Agriculture Sector") and by Mitacs through the Accelerate program, award IT40518.}

\conflictsofinterest{The authors declare no conflicts of interest.}

\begin{adjustwidth}{-\extralength}{0cm}

\reftitle{References}

\bibliography{ref}

@incollection{BASSO2019201,
title = {Chapter Four - Seasonal crop yield forecast: Methods, applications, and accuracies},
editor = {Donald L. Sparks},
series = {Advances in Agronomy},
publisher = {Academic Press},
volume = {154},
pages = {201-255},
year = {2019},
issn = {0065-2113},
doi = {https://doi.org/10.1016/bs.agron.2018.11.002},
url = {https://www.sciencedirect.com/science/article/pii/S0065211318300944},
author = {Bruno Basso and Lin Liu}
}

@article{
doi:10.1126/science.1204531,
author = {David B. Lobell  and Wolfram Schlenker  and Justin Costa-Roberts },
title = {Climate Trends and Global Crop Production Since 1980},
journal = {Science},
volume = {333},
number = {6042},
pages = {616-620},
year = {2011},
doi = {10.1126/science.1204531},
URL = {https://www.science.org/doi/abs/10.1126/science.1204531},
eprint = {https://www.science.org/doi/pdf/10.1126/science.1204531}}

@article{lesk2016influence,
  title={Influence of extreme weather disasters on global crop production},
  author={Lesk, Corey and Rowhani, Pedram and Ramankutty, Navin},
  journal={Nature},
  volume={529},
  number={7584},
  pages={84--87},
  year={2016},
  publisher={Nature Publishing Group UK London}
}

@article{shahhosseini2021coupling,
  title={Coupling machine learning and crop modeling improves crop yield prediction in the US Corn Belt},
  author={Shahhosseini, Mohsen and Hu, Guiping and Huber, Isaiah and Archontoulis, Sotirios V},
  journal={Scientific reports},
  volume={11},
  number={1},
  pages={1606},
  year={2021},
  publisher={Nature Publishing Group UK London}
}

@inproceedings{fan2022gnn,
  title={A GNN-RNN approach for harnessing geospatial and temporal information: application to crop yield prediction},
  author={Fan, Joshua and Bai, Junwen and Li, Zhiyun and Ortiz-Bobea, Ariel and Gomes, Carla P},
  booktitle={Proceedings of the AAAI conference on artificial intelligence},
  volume={36},
  number={11},
  pages={11873--11881},
  year={2022}
}

@article{khaki2020cnn,
  title={A CNN-RNN framework for crop yield prediction},
  author={Khaki, Saeed and Wang, Lizhi and Archontoulis, Sotirios V},
  journal={Frontiers in Plant Science},
  volume={10},
  pages={1750},
  year={2020},
  publisher={Frontiers Media SA}
}

@article{leng2019crop,
  title={Crop yield sensitivity of global major agricultural countries to droughts and the projected changes in the future},
  author={Leng, Guoyong and Hall, Jim},
  journal={Science of the Total Environment},
  volume={654},
  pages={811--821},
  year={2019},
  publisher={Elsevier}
}

@article{weiss2020remote,
  title={Remote sensing for agricultural applications: A meta-review},
  author={Weiss, Marie and Jacob, Fr{\'e}d{\'e}ric and Duveiller, Grgory},
  journal={Remote sensing of environment},
  volume={236},
  pages={111402},
  year={2020},
  publisher={Elsevier}
}

@incollection{tahir2023application,
  title={Application of unmanned aerial vehicles in precision agriculture},
  author={Tahir, Muhammad Naveed and Lan, Yubin and Zhang, Yali and Wenjiang, Huang and Wang, Yingkuan and Naqvi, Syed Muhammad Zaigham Abbas},
  booktitle={Precision agriculture},
  pages={55--70},
  year={2023},
  publisher={Elsevier}
}

@article{oikonomidis2023deep,
  title={Deep learning for crop yield prediction: a systematic literature review},
  author={Oikonomidis, Alexandros and Catal, Cagatay and Kassahun, Ayalew},
  journal={New Zealand Journal of Crop and Horticultural Science},
  volume={51},
  number={1},
  pages={1--26},
  year={2023},
  publisher={Taylor \& Francis}
}

@article{mupangwa2020evaluating,
  title={Evaluating machine learning algorithms for predicting maize yield under conservation agriculture in Eastern and Southern Africa},
  author={Mupangwa, W and Chipindu, L and Nyagumbo, I and Mkuhlani, S and Sisito, G},
  journal={SN Applied Sciences},
  volume={2},
  number={5},
  pages={952},
  year={2020},
  publisher={Springer}
}

@article{kumar2024survey,
  title={A survey on deep learning approaches for crop disease analysis in precision agriculture},
  author={Kumar, S Praveen and Rao, Y Raghavender},
  journal={Turkish Journal of Computer and Mathematics Education},
  volume={15},
  number={1},
  pages={242--253},
  year={2024},
  publisher={Ninety Nine Publication}
}

@article{sharma2023precision,
  title={Precision agriculture: Reviewing the advancements technologies and applications in precision agriculture for improved crop productivity and resource management},
  author={Sharma, Shikha},
  journal={Reviews In Food and Agriculture},
  volume={4},
  number={2},
  pages={45--49},
  year={2023}
}

@inproceedings{pandey2024enhancing,
  title={Enhancing Crop Yield Prediction Accuracy in Precision Agriculture: A Comparative Analysis},
  author={Pandey, Anil Kumar and Kumar, Sanjay and Agrawal, Deepika and Pandey, Sudhakar},
  booktitle={2024 IEEE 1st International Conference on Advances in Signal Processing, Power, Communication, and Computing (ASPCC)},
  pages={236--241},
  year={2024},
  organization={IEEE}
}

@inproceedings{qomariyah2024applying,
  title={Applying Random Forest for Optimal Crop Selection to Enhance Agricultural Decision-Making},
  author={Qomariyah, Nurul and Putra, Septafiansyah Dwi and Afifah, Dian Ayu and Supriyatna, Agiska Ria and Zuriati, Zuriati},
  booktitle={7th International Conference on Applied Engineering (ICAE 2024)},
  pages={66--77},
  year={2024},
  organization={Atlantis Press}
}

@article{nevavuori2019crop,
  title={Crop yield prediction with deep convolutional neural networks},
  author={Nevavuori, Petteri and Narra, Nathaniel and Lipping, Tarmo},
  journal={Computers and electronics in agriculture},
  volume={163},
  pages={104859},
  year={2019},
  publisher={Elsevier}
}

@article{sun2019county,
  title={County-level soybean yield prediction using deep CNN-LSTM model},
  author={Sun, Jie and Di, Liping and Sun, Ziheng and Shen, Yonglin and Lai, Zulong},
  journal={Sensors},
  volume={19},
  number={20},
  pages={4363},
  year={2019},
  publisher={MDPI}
}

@article{lin2020deepcropnet,
  title={DeepCropNet: a deep spatial-temporal learning framework for county-level corn yield estimation},
  author={Lin, Tao and Zhong, Renhai and Wang, Yudi and Xu, Jinfan and Jiang, Hao and Xu, Jialu and Ying, Yibin and Rodriguez, Luis and Ting, KC and Li, Haifeng},
  journal={Environmental research letters},
  volume={15},
  number={3},
  pages={034016},
  year={2020},
  publisher={IOP Publishing}
}

@article{albahli2022efficient,
  title={Efficient attention-based CNN network (EANet) for multi-class maize crop disease classification},
  author={Albahli, Saleh and Masood, Momina},
  journal={Frontiers in Plant Science},
  volume={13},
  pages={1003152},
  year={2022},
  publisher={Frontiers Media SA}
}

@article{dosovitskiy2020image,
  title={An image is worth 16x16 words: Transformers for image recognition at scale},
  author={Dosovitskiy, Alexey},
  journal={arXiv preprint arXiv:2010.11929},
  year={2020}
}

@article{szwarcman2024prithvi,
  title={Prithvi-eo-2.0: A versatile multi-temporal foundation model for earth observation applications},
  author={Szwarcman, Daniela and Roy, Sujit and Fraccaro, Paolo and G{\'\i}slason, {\TH}orsteinn El{\'\i} and Blumenstiel, Benedikt and Ghosal, Rinki and de Oliveira, Pedro Henrique and Almeida, Joao Lucas de Sousa and Sedona, Rocco and Kang, Yanghui and others},
  journal={arXiv preprint arXiv:2412.02732},
  year={2024}
}

@article{https://doi.org/10.2135/cropsci2003.1358,
author = {Angadi, S. V. and Cutforth, H. W. and McConkey, B. G. and Gan, Y.},
title = {Yield Adjustment by Canola Grown at Different Plant Populations under Semiarid Conditions},
journal = {Crop Science},
volume = {43},
number = {4},
pages = {1358-1366},
doi = {https://doi.org/10.2135/cropsci2003.1358},
url = {https://acsess.onlinelibrary.wiley.com/doi/abs/10.2135/cropsci2003.1358},
eprint = {https://acsess.onlinelibrary.wiley.com/doi/pdf/10.2135/cropsci2003.1358},
year = {2003}
}

@article{hodson2022root,
  title={Root mean square error (RMSE) or mean absolute error (MAE): When to use them or not},
  author={Hodson, Timothy O},
  journal={Geoscientific Model Development Discussions},
  volume={2022},
  pages={1--10},
  year={2022},
  publisher={G{\"o}ttingen, Germany}
}

@article{ozer1985correlation,
  title={Correlation and the coefficient of determination.},
  author={Ozer, Daniel J},
  journal={Psychological bulletin},
  volume={97},
  number={2},
  pages={307},
  year={1985},
  publisher={American Psychological Association}
}

@incollection{benesty2009pearson,
  title={Pearson correlation coefficient},
  author={Benesty, Jacob and Chen, Jingdong and Huang, Yiteng and Cohen, Israel},
  booktitle={Noise reduction in speech processing},
  pages={1--4},
  year={2009},
  publisher={Springer}
}

@article{nevavuori2020crop,
  title={Crop yield prediction using multitemporal UAV data and spatio-temporal deep learning models},
  author={Nevavuori, Petteri and Narra, Nathaniel and Linna, Petri and Lipping, Tarmo},
  journal={Remote sensing},
  volume={12},
  number={23},
  pages={4000},
  year={2020},
  publisher={MDPI}
}

@article{GAVAHI2021115511,
title = {DeepYield: A combined convolutional neural network with long short-term memory for crop yield forecasting},
journal = {Expert Systems with Applications},
volume = {184},
pages = {115511},
year = {2021},
issn = {0957-4174},
doi = {https://doi.org/10.1016/j.eswa.2021.115511},
url = {https://www.sciencedirect.com/science/article/pii/S0957417421009210},
author = {Keyhan Gavahi and Peyman Abbaszadeh and Hamid Moradkhani}
}

@article{holzman2021relationship,
  title={Relationship between TIR and NIR-SWIR as indicator of vegetation water availability},
  author={Holzman, Mauro Ezequiel and Rivas, Ra{\'u}l Eduardo and Bayala, Mart{\'\i}n Ignacio},
  journal={Remote Sensing},
  volume={13},
  number={17},
  pages={3371},
  year={2021},
  publisher={MDPI}
}

@article{wang2008sensitivity,
  title={Sensitivity studies of the moisture effects on MODIS SWIR reflectance and vegetation water indices},
  author={Wang, Lingli and Qu, John J and Hao, Xianjun and Zhu, Qingping},
  journal={International Journal of Remote Sensing},
  volume={29},
  number={24},
  pages={7065--7075},
  year={2008},
  publisher={Taylor \& Francis}
}

@inproceedings{hunt2007remote,
  title={Remote sensing of vegetation water content using shortwave infrared reflectances},
  author={Hunt Jr, E Raymond and Yilmaz, M Tugrul},
  booktitle={Remote Sensing and Modeling of Ecosystems for Sustainability IV},
  volume={6679},
  pages={15--22},
  year={2007},
  organization={SPIE}
}

@Article{agronomy14020327,
AUTHOR = {Barman, Utpal and Sarma, Parismita and Rahman, Mirzanur and Deka, Vaskar and Lahkar, Swati and Sharma, Vaishali and Saikia, Manob Jyoti},
TITLE = {ViT-SmartAgri: Vision Transformer and Smartphone-Based Plant Disease Detection for Smart Agriculture},
JOURNAL = {Agronomy},
VOLUME = {14},
YEAR = {2024},
NUMBER = {2},
ARTICLE-NUMBER = {327},
URL = {https://www.mdpi.com/2073-4395/14/2/327},
ISSN = {2073-4395},
DOI = {10.3390/agronomy14020327}
}

@misc{mehdipour2025visiontransformersprecisionagriculture,
      title={Vision Transformers in Precision Agriculture: A Comprehensive Survey}, 
      author={Saber Mehdipour and Seyed Abolghasem Mirroshandel and Seyed Amirhossein Tabatabaei},
      year={2025},
      eprint={2504.21706},
      archivePrefix={arXiv},
      primaryClass={cs.CV},
      url={https://arxiv.org/abs/2504.21706}, 
}

@Article{s23156949,
AUTHOR = {Parez, Sana and Dilshad, Naqqash and Alghamdi, Norah Saleh and Alanazi, Turki M. and Lee, Jong Weon},
TITLE = {Visual Intelligence in Precision Agriculture: Exploring Plant Disease Detection via Efficient Vision Transformers},
JOURNAL = {Sensors},
VOLUME = {23},
YEAR = {2023},
NUMBER = {15},
ARTICLE-NUMBER = {6949},
URL = {https://www.mdpi.com/1424-8220/23/15/6949},
PubMedID = {37571732},
ISSN = {1424-8220},
DOI = {10.3390/s23156949}
}

@techreport{UN_WPP_2024,
  author = {{United Nations, Department of Economic and Social Affairs, Population Division}},
  title = {{World Population Prospects 2024: Summary of Results}},
  year = {2024},
  institution = {{United Nations}},
  address = {New York},
  note = {Report No. UN DESA/POP/2024/TR/NO. 9. Available at \url{https://population.un.org/wpp/}},
  url = {https://population.un.org/wpp/}
}

\PublishersNote{}
\end{adjustwidth}
\end{document}